\pdfoutput=1

\documentclass[11pt]{article}

\usepackage[final]{acl}

\usepackage{times}
\usepackage{latexsym}

\usepackage[T1]{fontenc}

\usepackage[utf8]{inputenc}

\usepackage{microtype}

\usepackage{inconsolata}

\usepackage{multirow}
\usepackage{booktabs}
\usepackage{graphicx}
\usepackage{amsmath}
\usepackage{amsfonts}
\usepackage{makecell}
\usepackage{hyperref}
\usepackage{stfloats}
\usepackage{tablefootnote}
\usepackage{algorithm}
\usepackage{algorithmic}

\usepackage{color}
\usepackage{soul}
\usepackage{makecell}
\usepackage{multicol}
\usepackage{supertabular}
\usepackage{listings}
\usepackage{arydshln}
\usepackage{tocloft}
\usepackage{adjustbox}

\definecolor{lightblue}{rgb}{.8,.8,1}

\title{ExpeTrans: LLMs Are Experiential Transfer Learners}

\author{Jinglong Gao\quad
Xiao Ding\footnotemark[1]\quad
Lingxiao Zou\quad
Bibo Cai \quad Bing Qin \quad Ting Liu \\
\normalsize{Research Center for Social Computing and Interactive Robotics}\\[-.05cm]
\normalsize{Harbin Institute of Technology, China}\\[-.05cm]
{\small\tt\{jlgao, xding, lxzou, bbcai, qinb, tliu\}@ir.hit.edu.cn}\\[-.05cm]}

\begin{document}
\maketitle
\begin{abstract}

\renewcommand{\thefootnote}{\fnsymbol{footnote}}
\footnotetext[1]{Corresponding Author}

Recent studies provide large language models (LLMs) with textual task-solving experiences via prompts to improve their performance.
However, previous methods rely on substantial human labor or time to gather such experiences for each task, which is impractical given the growing variety of task types in user queries to LLMs.
To address this issue, we design an autonomous experience transfer framework to explore whether LLMs can mimic human cognitive intelligence to autonomously transfer experience from existing source tasks to newly encountered target tasks. This not only allows the acquisition of experience without extensive costs of previous methods, but also offers a novel path for the generalization of LLMs.
Experimental results on 13 datasets demonstrate that our framework effectively improves the performance of LLMs. Furthermore, we provide a detailed analysis of each module in the framework.

\end{abstract}

\section{Introduction}
\begin{figure}[t]
  \centering
\includegraphics[width=1\linewidth,trim={1cm 1cm 1cm 1cm},clip]{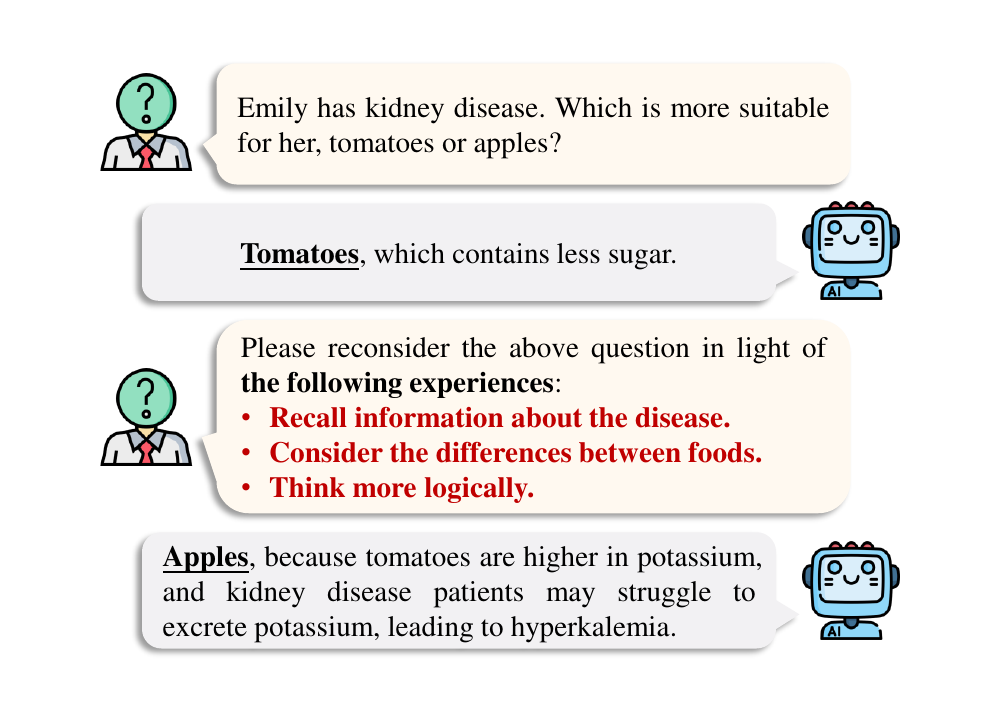}
  \caption{An example of LLMs inference guided by textual experience.}
  \label{fig:intro}
\end{figure}

Recently, LLMs like ChatGPT have achieved remarkable performance in various NLP tasks \citep{ye2023comprehensive}. However, many NLP tasks still cannot be effectively addressed by LLMs \citep{chang2023survey,chu-etal-2024-timebench}. Previous research finds that this issue can be mitigated by using prompts to provide LLMs with textual task-solving experience during their inference stage (as shown in Figure~\ref{fig:intro}). The experiences they utilize are textual descriptions of task-solving strategies, steps, insights, etc.

Previous studies mainly focus on how to obtain such experience. Initially, some studies manually write experience for each task \citep{wei2022chain,kong2023better}. Later, researchers leverage LLMs to automatically summarize experience from manually labeled datasets of each task \citep{zhao2023expel,chen2024grimoire}. Besides, for each user query, \citet{gao-etal-2024-self-evolving} utilizes LLMs to generate lots of similar queries and assigns labels to them to obtain a pseudo-dataset (taking 3 to 8 minutes per user query), from which the experience is then summarized.
However, whenever these methods encounter a new task, they require substantial human labor or time to gain experiences for that task.
Given the growing variety of new task types in user queries to LLMs, these methods are becoming increasingly impractical.

In contrast, humans can transfer experiences from existing source tasks to newly encountered target tasks, thus acquiring experiences without the extensive cost of learning from scratch. According to Structure-Mapping Theory (SMT) \citep{GENTNER1983155}, humans primarily consider two aspects when assessing task transferability and selecting source tasks: 1) Task Function similarity: experience transfer often occurs between tasks that have similar goals; 2) Task Process similarity: experience transfer is likely between tasks that share similar execution steps. Once the source tasks are identified, humans can flexibly adapt the source task experience to the target task.
Inspired by this, we want to explore whether LLMs can mimic the above process. This could not only avoid the substantial costs of previous methods but also offer a novel path for the generalization of LLMs.

\begin{figure*}[t]
  \centering
  \includegraphics[width=\linewidth,trim={0.5cm 0.5cm 0.5cm 0.5cm},clip]{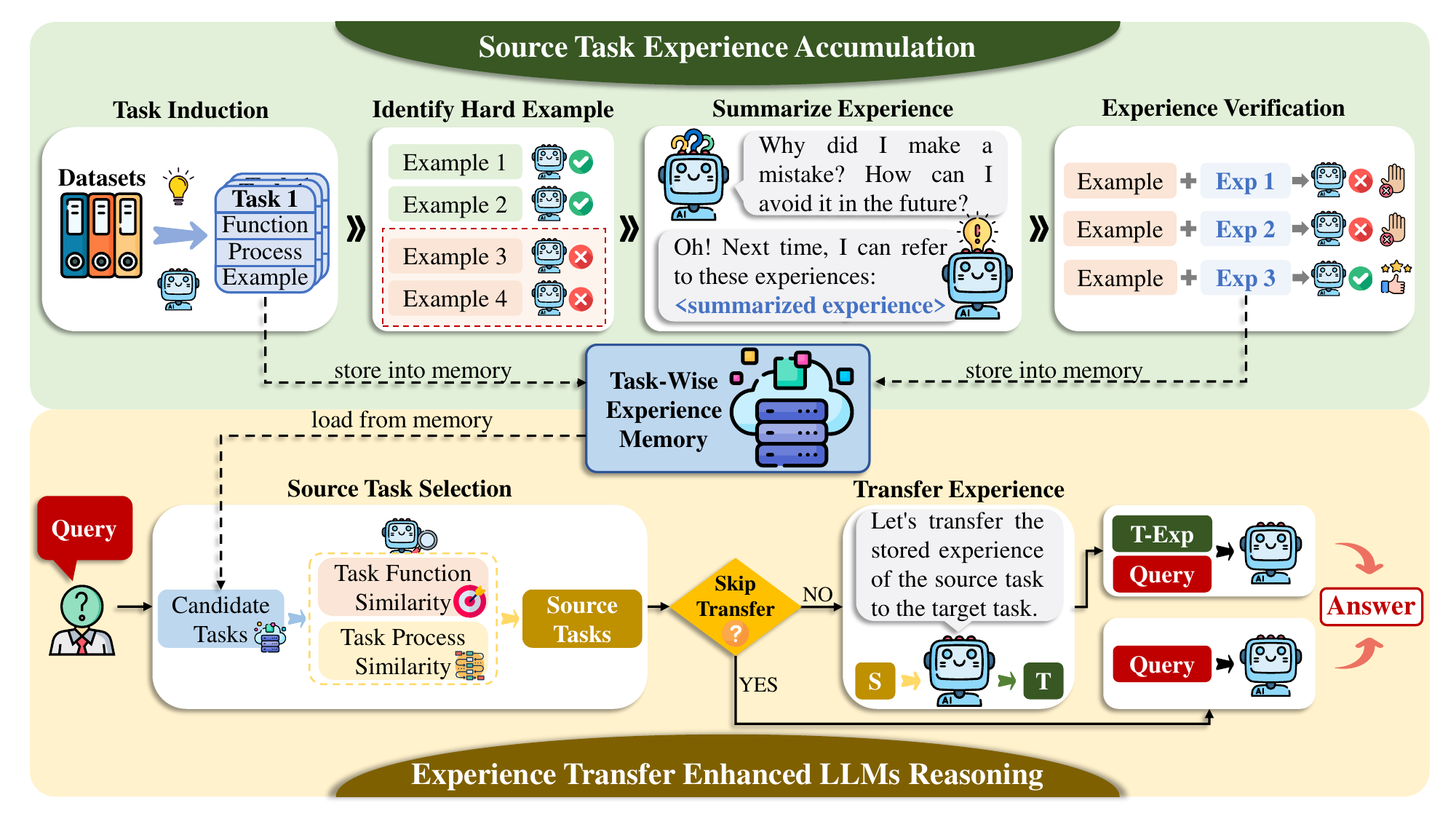}
  \caption{The framework of our proposed ExpeTrans.}
  \label{fig:framework}
\end{figure*}

To achieve this, we propose an autonomous experience transfer framework called \textbf{ExpeTrans}, which consists of a task-wise experience memory and multiple experience-processing modules based on ChatGPT.
Our framework begins with the source task experience accumulation. It autonomously analyzes existing labeled datasets to summarize various tasks, including their functions (i.e., goals), processes (i.e., steps), and experiences. These are then stored in memory as candidate source tasks.
Subsequently, our framework processes each user query. It first identifies the function and process of the target task to which the user query belongs. Then, based on the similarities in task function and process, several suitable source tasks are selected from memory, and their experiences are flexibly transferred to the target task. Finally, our framework combines the transferred experience to respond to the user.

Our contributions are summarized as follows:
\begin{enumerate}
    \item We study whether LLMs can mimic human transfer learning, which not only allows the acquisition of experience without extensive costs of previous methods, but also offers a novel path for the generalization of LLMs.
    \item We design a LLMs-based framework ExpeTrans, performing cognitively inspired autonomous experience acquisition, accumulation, and transfer.
    \item Experiments on 13 widely used NLP datasets show that our framework can effectively transfer experience across tasks. Additionally, we provide a detailed analysis of each step of the experience transfer, as well as the effect of the task transferability level.
\end{enumerate}

\section{Methodology}

Figure~\ref{fig:framework} shows the framework of our proposed ExpeTrans, which consists of a task-wise experience memory and multiple experience-processing modules based on ChatGPT.

Our framework starts by autonomously accumulating source task experience. It analyzes existing labeled NLP datasets, summarizing various tasks with their functions (i.e., goals), processes (i.e., steps), and experiences, which are then stored in memory as candidate source tasks.
Subsequently, our framework autonomously handles user queries.
For each query, it first identifies the function and process of the target task. It then selects several source tasks from memory based on task function and process similarities. The experiences from these tasks are flexibly transferred to the target task. Finally, it responds to the user query with the transferred experience.
The prompts and examples of our framework are presented in Appendix~\ref{appendix:prompts}.

\subsection{Task-Wise Experience Memory}
\label{sec:memory}

We establish a task-wise experience memory to store tasks and experiences gained during the source task experience accumulation phase (\S\ref{sec:stea}).

Specifically, this memory stores three fields for each task: 1) Task Function, which refers to the goal that the task aims to achieve; 2) Task Process, which outlines the steps to infer the task; and 3) Task Experience, multiple insights that help solve the task more effectively and avoid errors.
All of this content is autonomously summarized by LLMs.
Appendix~\ref{sec:example_memory} provides examples of the memory.

\subsection{Source Task Experience Accumulation}
\label{sec:stea}
At this stage, we guide LLMs to autonomously summarize various tasks and their experiences from existing labeled datasets, which serve as candidate source tasks for experience transfer.

\subsubsection{Task Granularity and Data Utilization}
\label{sec:tgdu}
As stated in \S\ref{related:EL}, previous studies also summarized experiences from labeled datasets. However, they treated the entire dataset as a single task and only selected a dozen examples from it to summarize the experience. This setting is unsuitable for experience transfer:
1) examples within the same dataset may vary in terms of specific goals, reasoning steps, and scenarios, making them suitable as distinct source tasks for transfer to different target tasks; 2) the utilization of existing data is too low, leading to insufficient transfer.
In contrast, we utilize all examples from the dataset, treating each example as a separate source task to perform comprehensive and fine-grained transfer of experience.

\subsubsection{Experience Accumulation Process}
\label{sec:eap}
Our framework processes each example from the labeled datasets in parallel. For each example, this phase is divided into four steps:
1) ChatGPT utilizes Prompt~\ref{pt:p1} and Prompt~\ref{pt:p2} to generate the function and process of the task to which the input example belongs. These will be used in \S\ref{sec:sts} to select source tasks;
2) according to previous studies \citep{yang-etal-2023-failures, chen2024grimoire}, the experience summarized from hard examples is much more effective. Thus, we use ChatGPT to answer the example with the Zero-shot-CoT \citep{zscot}, i.e., adding ``Let's think step by step.'' after the original question. If the answer is incorrect, the example is regarded as a hard example; otherwise, no further steps are taken;
3) we use Prompt~\ref{pt:p3} to guide ChatGPT in reflecting on its incorrect answer for the hard example in the previous step, summarizing experiences about how to solve the task and avoid errors;
4) for the experiences, we use Prompt~\ref{pt:p4} to verify their effectiveness, retaining only those that enable ChatGPT to correctly respond to the current hard example.

Finally, the task functions, processes, and experiences obtained from each example are individually stored in our memory (\S\ref{sec:memory}). Please note that for non-hard examples, their experience is stored as empty and is used to determine whether experience transfer is needed (detailed in \S\ref{sec:sts}).

\subsection{Experience Transfer Enhanced LLMs Reasoning}
\label{sec:allinference}
After the experience accumulation phase, our framework begins autonomously processing user queries. For each user query, it selects appropriate source tasks and transfers the source task experience to the target task to which the user query belongs. Finally, it uses the transferred experience to respond to the user query.

\subsubsection{Source Task Selection}
\label{sec:sts}

According to SMT, humans primarily consider the similarity in task functions and task processes when assessing transferability across tasks. Inspired by this, we also take these two aspects into account when selecting source tasks for the target task.

This stage consists of four steps: 1) ChatGPT employs Prompt~\ref{pt:p1} and Prompt~\ref{pt:p2} to generate the function and processe of the target task corresponding to the user query; 2) all candidate tasks in memory are ranked according to their BM25-score \citep{robertson1995okapi} with respect to the target task in terms of task function and task process, respectively; 3) the harmonic mean of these two rankings is calculated, encouraging source tasks that align closely with the target task in both aspects; 4) we use Prompt~\ref{pt:p5} to guide ChatGPT in selecting the source task from the top $K$ candidates in the harmonic mean ranking.

\subsubsection{Transfer or Skip Transfer}
\label{sec:skiptrans}
Not all queries require transfer, and inappropriate transfer may not only waste time but also potentially reduce the performance of LLMs.

Firstly, there may not be suitable source tasks in memory. Therefore, in Prompt~\ref{pt:p5} of \S\ref{sec:sts}, we allow ChatGPT to return an empty list, indicating that no suitable source task was found.

Secondly, the user query may be simple, and LLMs can answer it correctly on their own. To accommodate this, in \S\ref{sec:eap}, we store both hard and non-hard examples in memory, with the latter containing no stored experience. This means that in \S\ref{sec:sts}, if our framework considers the most suitable source tasks to be non-hard examples, no experience will be available for transfer.

In both cases above, we let ChatGPT directly respond to the user query using Zero-shot-CoT.

\subsubsection{Transfer Experience}
\label{sec:ttansferee}
After selecting the source tasks, we transfer the experience from multiple source tasks to the target task. To ensure a quick response, we process each source task in parallel.

Specifically, we guide ChatGPT with Prompt~\ref{pt:p6} to adapt the experience of each source task to the target task. We only provide ChatGPT with the task functions and processes of the source and target tasks, without providing any examples or queries. We find that such prompt better maintains the generalizability of the transferred experience.

\subsubsection{Reasoning with Experience}
\label{sec:rwe}
Finally, our framework responds to the user query with the transferred experience.

In practice, we find that too much experience is costly but does not effectively improve model performance. Therefore, we employ BM25 to select the top $M$ insights from the transferred experience that are most relevant to the query. Subsequently, we use Prompt~\ref{pt:p7} to guide ChatGPT in responding to the user with selected insights.

\section{Experiments}
\subsection{Datasets and Evaluation Metrics}
\label{datasets}
We conduct experiments on four major categories that include 13 datasets (Appendix~\ref{sec:example_dataset} shows examples).
The abbreviations for each dataset are highlighted in \textbf{bold}.
Firstly, the \textbf{\emph{Event Relation Understand}} category includes four datasets from the MAVEN-ERE \citep{wang-etal-2022-maven} benchmark: 1) \textbf{Time} Relation Classification, 2) \textbf{Causal} Relation Identification, 3) Event \textbf{Coref}erence Resolution, and 4) \textbf{Sub-E}vent Recognition.
Secondly, the \textbf{\emph{Textual Reasoning}} category includes: 1) bAbI-\textbf{Induc}tion and 2) bAbI-\textbf{Deduc}tion \citep{weston2015towards}: contains a series of problems requiring induction and deduction reasoning, respectively; 3) \textbf{Conj} \citep{saha-etal-2020-conjnli}: tests natural language inference over conjunctive sentences; 4) \textbf{SNLI} \citep{bowman-etal-2015-large}: mainly focuses on natural language inference in everyday life scenarios.
Thirdly, the \textbf{\emph{Sentiment Analysis}} category includes:
1) \textbf{Twitter}17 \citep{rosenthal-etal-2017-semeval}: Sentiment classification on Twitter texts; 2) \textbf{Stance}16 \citep{mohammad-etal-2016-semeval}: Detects the stance of a text towards a specific topic; 3) \textbf{Lap}14 \citep{pontiki-etal-2014-semeval}: Classifies the sentiment of a text towards a particular entity or aspect.
Finally, the \textbf{\emph{Discrimination-Safe Response (DSR)}} category includes two datasets from the BBQ-Lite \citep{srivastava2023beyond} benchmark: 1) \textbf{Age} and 2) \textbf{Dis}ability, where the model is required to avoid selecting biased options related to age or disability discrimination, respectively.

We randomly select 500 examples from each dataset and mix them as the test data, which contains 6,500 examples.
In addition to testing, all of these examples are also used for our experience accumulation.
We adopt a cross-validation setting to test our framework. Specifically, each test example is not allowed to select source tasks summarized from the same dataset it belongs to, i.e., the experiences are all transferred from other datasets.
We employ Accuracy (Acc) as the evaluation metric and report the average Acc of three rounds of predictions to reduce randomness. Besides, ``\textit{Avg.}'' denotes the average Acc across 13 datasets.

\begin{table*}[!t]
\small
\centering
\setlength{\tabcolsep}{3.15pt}
\begin{tabular}{lcccccccccccccc}
\toprule
\multirow{2}{*}{\textbf{Method}} & \multicolumn{4}{c}{\textbf{Event Relation Understand}}& \multicolumn{4}{c}{\textbf{Textual Reasoning}}& \multicolumn{3}{c}{\textbf{Sentiment Analysis}}& \multicolumn{2}{c}{\textbf{DSR}} & \multirow{2}{*}{\textit{\textbf{Avg.}}}\\
\cmidrule(lr){2-5} \cmidrule(lr){6-9} \cmidrule(lr){10-12} \cmidrule(lr){13-14}
& \textbf{Time} & \textbf{Causal} & \textbf{Coref} & \textbf{Sub-E} & \textbf{Induc} & \textbf{Deduc} & \textbf{Conj}&\textbf{SNLI}&\textbf{Twitter}&\textbf{Stance}&\textbf{Lap}&\textbf{Age}&\textbf{Dis} &  \\
\midrule
\textbf{Zero-shot} & 37.3 & 24.8 & 11.4 & 30.9 & 52.4 & 70.3 & 59.6 & 62.0 & 49.3 & 50.0 & 71.2 & 58.0 & 61.4 & 49.1\\
\textbf{ZS-CoT} & 38.0 & 24.4 & 30.8 & 43.3 & 53.6 & \underline{76.2} & 55.6 & 63.6 & 53.3 & 56.3 & 74.4 & 69.6 & 74.5 & 54.9\\
\textbf{Self-EXP} & 32.2 & 19.1 & 32.3 & 40.6 & 51.8 & 75.5 & 52.4 & 62.0 & 54.8 & 53.3 & 68.2 & 57.8 & 75.8 & 52.0\\
\textbf{SEGPT-T} & 39.2 & 25.1 & \underline{35.9} & 47.9 & 53.0 & 75.3 & \underline{60.0} & 62.1 & \underline{54.9} & \underline{56.8} & \underline{76.0} & 68.3 & 73.6 & 56.1\\
\textbf{Cross-ICL} & \underline{42.0} & \underline{39.6} & 30.4 & \underline{58.8} & \underline{53.9} & 58.4 & 57.8 & \textbf{73.6} & 53.2 & 52.0 & 75.0 & \textbf{78.7} & \textbf{84.2} & \underline{58.2}\\
\textbf{Cross-ICL*} & 37.3 & \textbf{40.6} & 26.0 & 45.8 & 53.5 & 66.1 & 54.0 & 59.6 & 52.9 & 52.9 & 72.0 & 62.0 & 63.5 & 52.8\\
\textbf{Ours} & \textbf{43.4} & 33.0 & \textbf{57.6} & \textbf{60.7} & \textbf{57.9} & \textbf{81.2} & \textbf{63.7} & \underline{70.4} & \textbf{60.8} & \textbf{60.8} & \textbf{80.8} & \underline{76.2} & \underline{82.4} & \textbf{63.8}\\

\bottomrule
\end{tabular}
\caption{\label{tab:main}Experimental results (\%) on the mixture of 13 datasets. \textbf{Bold} and \underline{Underlined} numbers represent the 1st and the 2nd best performance on each dataset.
``\textit{Avg.}'' denotes the mean accuracy across datasets for each method.}
\end{table*}

\begin{table*}[!t]
\small
\centering
\setlength{\tabcolsep}{3.6pt}
\begin{tabular}{lcccccccccccccc}
\toprule
\multirow{2}{*}{\textbf{Method}} & \multicolumn{4}{c}{\textbf{Event Relation Understand}}& \multicolumn{4}{c}{\textbf{Textual Reasoning}}& \multicolumn{3}{c}{\textbf{Sentiment Analysis}}& \multicolumn{2}{c}{\textbf{DSR}} & \multirow{2}{*}{\textit{\textbf{Avg.}}}\\
\cmidrule(lr){2-5} \cmidrule(lr){6-9} \cmidrule(lr){10-12} \cmidrule(lr){13-14}
& \textbf{Time} & \textbf{Causal} & \textbf{Coref} & \textbf{Sub-E} & \textbf{Induc} & \textbf{Deduc} & \textbf{Conj}&\textbf{SNLI}&\textbf{Twitter}&\textbf{Stance}&\textbf{Lap}&\textbf{Age}&\textbf{Dis} &  \\
\midrule
\textbf{Normal} & \textbf{43.4} & \textbf{33.0} & \textbf{57.6} & \textbf{60.7} & \textbf{57.9} & \textbf{81.2} & \textbf{63.7} & \textbf{70.4} & \textbf{60.8} & \textbf{60.8} & \textbf{80.8} & \textbf{76.2} & \textbf{82.4} & \textbf{63.8}\\
\textbf{Low} & \underline{40.3} & \underline{30.9} & \underline{44.7} & \underline{54.8} & \underline{57.8} & \underline{79.3} & \underline{62.5} & \underline{68.7} & \underline{60.7} & \underline{58.2} & \underline{78.5} & \underline{73.1} & \underline{76.2} & \underline{60.4}\\
\textbf{Minimal} & 37.0 & 25.6 & 36.6 & 43.2 & 53.4 & 72.9 & 60.1 & 64.0 & 56.3 & 55.1 & 75.3 & 69.8 & 73.1 & 55.5\\

\bottomrule
\end{tabular}
\caption{\label{tab:othertranslevel}Performance (\%) of our framework based on GPT-3.5 under different levels of task transferability.}
\end{table*}

\subsection{Parameters Setting}
Our experiments are mainly based on \textbf{\texttt{gpt-3.5-turbo-0125} (GPT-3.5)}. We also discuss the performance on \textbf{six smaller or larger LLMs in Appendix~\ref{app:smallllm}}, including \texttt{gpt-4o-2024-05-13} (GPT-4).
We manually select hyperparameters.
\texttt{temperature} is the default value 1.
For the Prompt~\ref{pt:p4} in \S\ref{sec:eap}, we execute it in parallel 3 times, retaining only the experiences that pass all 3 checks.
$K$ in \S\ref{sec:sts} is set to 10, and $M$ in \S\ref{sec:rwe} is set to 5, with their effects discussed in \S\ref{app:candi_num} and \S\ref{sec:num_M}, respectively.

\subsection{Baselines}
We employ the following baseline methods in our experiments:
1) \textbf{Zero-shot}, ChatGPT performs zero-shot reasoning for the query;
2) \textbf{Zero-shot-CoT (ZS-CoT)}, add ``Let's think step by step'' after the query to let ChatGPT think step by step;
3) \textbf{Self-EXP} \citep{gao-etal-2024-self-evolving}, which prompts ChatGPT to directly generate experience for the query, and then think the query with the generated experience;
4)~\textbf{SEGPT-T}ransfer, the modification of SEGPT \citep{gao-etal-2024-self-evolving} that first summarizes experience from our test data instead of its noisy pseudo-dataset, then performs its auxiliary module to rewrite common insights for thinking the query. Same as ours, it is tested in the cross-validation setting;
5) \textbf{Cross-ICL}, performs cross-task ICL inspired by \citet{chatterjee-etal-2024-language}. For each query, 
we select 5 examples from the other 12 datasets to serve as ICL demonstrations in a CoT style for the query (detailed in Appendix~\ref{app:cross-icl});
6) \textbf{Cross-ICL*}, the same as Cross-ICL but selects demonstrations only from datasets within the other 3 categories, rather than all 12 datasets.
Our framework is also compared with other ICL variants in Appendix~\ref{evse}.

\subsection{Main Results}
\label{sec:main-exp-results}
Table~\ref{tab:main} shows our experimental results on the mixture of 13 datasets. We find that:

Firstly, our framework consistently outperforms the baseline methods.
This is mainly because our framework effectively performs experience transfer, utilizing experience from known tasks to solve new tasks. We also discuss speed and cost in \S\ref{append:speed} and present the case study in Appendix~\ref{app:casestudy}.

Secondly, Self-EXP does not effectively improve the average accuracy. This is because the quality of the experience generated by LLMs is unreliable, often containing insights that LLMs have already mastered \citep{gao-etal-2024-self-evolving}. In contrast, our method summarizes and verifies experiences using labeled data to obtain reliable experience for the transfer process.

Thirdly, SEGPT-T only achieves limited improvement for LLMs. This is because SEGPT primarily focuses on acquiring experience through generating pseudo-datasets, rather than experience transfer. Its auxiliary module simply captures shared common insights across multiple tasks, which are often too generic or trivial.
In contrast, our work focuses on experience transfer, achieved through careful design to ensure thorough and reliable transfer of experience.

Besides, compared to ZS-CoT, Cross-ICL* reduces the average performance.
It is because it relies on specific examples and is highly sensitive to format differences between tasks.
The improvement of Cross-ICL is due to the similar formats within our same-category datasets, which are not guaranteed in real-world queries.
In contrast, as shown in Table~\ref{tab:othertranslevel}, our framework can effectively improve the performance in the same test setting as Cross-ICL*.
This is because we leverage experience that can generalize well across tasks.

\subsection{The Impact of Task Transferability}
\label{effect:impact_translity}

As shown in Table~\ref{tab:othertranslevel}, we evaluate our framework under different levels of task transferability: 1)~\textbf{Normal}, each test example is only allowed to select source tasks from other datasets (same as stated in \S\ref{datasets}); 2) \textbf{Low}, each test example is only allowed to select source tasks from other three major categories (same as CrossICL*); 3) \textbf{Minimal}, based on the \textbf{Low} setting, we select the bottom tasks rather than the top tasks from the ranking (Step 4 in \S\ref{sec:sts}) as candidate source tasks. We find that:

First, our framework is still effective under the \textbf{Low} setting, surpassing Zero-shot by 11.3\% (\textit{Avg.}). This indicates that very similar source tasks are beneficial but not essential for our framework. \textbf{Our framework remains effective when transferring between less similar tasks.} 
Moreover, there is already a vast number of annotated datasets available within the NLP community, so the primary goal of our framework is to make better use of these existing datasets, rather than annotate additional datasets as source tasks. Besides, we also analyze the intra-dataset transfer in Appendix~\ref{app:samesetexp}.

Secondly, in the extreme and harsh \textbf{Minimal} setting, our framework is still comparable to ZS-CoT and does not hurt the origin performance of the LLM. This is because our skip strategy in \S\ref{sec:skiptrans} allows for dynamically skipping experience transfer for unsuitable queries.

\subsection{Speed and Prompt Cost}
\label{append:speed}

\begin{table*}[!t]
\small
\centering
\begin{tabular}{lccccc}
\toprule
\multirow{2}{*}{\textbf{Method}} & \multicolumn{3}{c}{\textbf{Tokens}} & \multirow{2}{*}{\textbf{Time (s)}} & \multirow{2}{*}{\textbf{Parallel Execution}} \\
\cmidrule(lr){2-4}
& \textbf{Prompt}& \textbf{Completion}& \textbf{Total} & &\\
\midrule
\rowcolor{gray!20} 
\multicolumn{6}{c}{\textbf{\emph{Preparation Phase}}}\\
\textbf{Ours}\\
\textbf{ - Experience Accumulation} \S\ref{sec:stea} & 9,257 & 2,773 & 12,030 & 7.45 & yes\\
\midrule
\rowcolor{gray!20} 
\multicolumn{6}{c}{\textbf{\emph{Inference Phase}}}\\
\textbf{Ours}\\
\textbf{ - Source Task Selection} \S\ref{sec:sts} & 1,319 & 22 & 1,341 & 1.57  &   no   \\
\textbf{ - Transfer Experience} \S\ref{sec:ttansferee} & 589 &130 &719 & 2.78 & yes\\
\textbf{ - Reasoning with Experience} \S\ref{sec:rwe} & 152 & 139 & 291 & 4.16 & no\\
\textbf{ - Total Inference} & 2,060 & 291 & 2,351 & 8.51 & no\\
\textbf{Zero-shot} & 114&38 & 152& 2.17 & no \\
\textbf{Zero-shot-CoT} & 122 & 112 & 234 & 4.09 & no\\
\textbf{Self-EXP} & 390 &547 & 937 & 6.35 & no \\
\textbf{Cross-ICL} & 1,352   &151 &1,503 & 6.52 & no \\
\textbf{5-shot ICL (with CoT)} & 1,343   &147 &1,490 & 6.38 & no \\
\textbf{SEGPT} & 17,429 & 4,340 & 21,769 &  262 & no \\

\bottomrule
\end{tabular}
\caption{\label{tab:speed}The speed and prompt costs based on GPT-3.5, averaged by the number of examples.}
\end{table*}

As shown in Table~\ref{tab:speed}, we present the prompt consumption and execution speed of our framework, averaged based on the number of examples. It can be observed that it takes only 8.51 seconds from receiving a user query to providing a response.
In comparison, Zero-shot, Zero-shot-CoT, and Self-EXP, while superior in terms of speed and cost to our method, exhibit poor performance. On the other hand, Cross-ICL demonstrates limited cross-task generalization capabilities (as indicated by Cross-ICL* in \S\ref{sec:main-exp-results}) and has inference costs and speed comparable to our method. Additionally, compared to our method, SEGPT is inferior in both cost and speed. In the inference phase, although our method uses more tokens than Zero-shot-CoT, \textbf{our method primarily focuses on prompt tokens}. Typically, the API price of prompt tokens is four to six times lower than that of complete tokens, so it does not result in a significant economic burden. In short, \textbf{our cost distribution is similar to the ICL techniques} (as ``5-shot ICL (with CoT)'' in Table~\ref{tab:speed}, where five CoT demonstrations are provided for each query in ICL).

Besides, \textbf{our framework achieves acceptable and sustainable economic efficiency.} On one hand, as mentioned in \S\ref{effect: num-check}, even if the number of samples used in the experience accumulation phase is reduced to one-tenth, our method is still effective. On the other hand, previous methods require labeling data for each new task, leading to continuously increasing costs. In contrast, our method can reuse initial existing experience for new tasks. As more tasks are solved, the cost of acquiring initial experience will be further distributed, allowing our average cost to continue decreasing.

\subsection{Analysis of the Source Task Experience Accumulation}

\paragraph{Effect of Source Task Granularity}
\label{effect:gst}

As shown in Table~\ref{tab:src_gran}, we discuss the impact of the granularity of the source task for our framework, specifically how many examples are considered together as one source task for source task selection (\S\ref{sec:sts}). There are three granularities: 1) \textbf{Ours}, as stated in \S\ref{sec:tgdu}, one source task only contains one example; 2) \textbf{-w/ Seg}, where we employ the ChatGPT-based module of \citet{gao-etal-2024-self-evolving} that automatically divides the dataset into tasks containing several dozen examples; 3) \textbf{-w/ Overall}, all examples in a dataset are treated as a single source task, which aligns with most previous methods. We find that:

Firstly, \textbf{-w/ Overall} significantly reduces the performance. This is mainly because even in a single dataset, the examples can differ in terms of specific goals, reasoning steps and scenarios, making them more suitable for different target tasks.

Secondly, despite dividing the dataset, \textbf{-w/ Seg} also reduces our performance. This is due to its division may not be effective for experience transfer, and finding a good division is challenging. Besides, samples from the same task may contain different scenarios, making it unsuitable to transfer all of them to a single target task.

\begin{table*}[!t]
\small
\centering
\setlength{\tabcolsep}{3pt}
\begin{tabular}{lcccccccccccccc}
\toprule
\multirow{2}{*}{\textbf{Method}} & \multicolumn{4}{c}{\textbf{Event Relation Understand}}& \multicolumn{4}{c}{\textbf{Textual Reasoning}}& \multicolumn{3}{c}{\textbf{Sentiment Analysis}}& \multicolumn{2}{c}{\textbf{DSR}} & \multirow{2}{*}{\textit{\textbf{Avg.}}}\\
\cmidrule(lr){2-5} \cmidrule(lr){6-9} \cmidrule(lr){10-12} \cmidrule(lr){13-14}
& \textbf{Time} & \textbf{Causal} & \textbf{Coref} & \textbf{Sub-E} & \textbf{Induc} & \textbf{Deduc} & \textbf{Conj}&\textbf{SNLI}&\textbf{Twitter}&\textbf{Stance}&\textbf{Lap}&\textbf{Age}&\textbf{Dis} &  \\
\midrule
\textbf{Ours} & \textbf{43.4} & \textbf{33.0} & \textbf{57.6} & \textbf{60.7} & \textbf{57.9} & \textbf{81.2} & \textbf{63.7} & \textbf{70.4} & \textbf{60.8} & \textbf{60.8} & \textbf{80.8} & \textbf{76.2} & \textbf{82.4} & \textbf{63.8}\\
\textbf{-w/ Seg} & \underline{39.8} & \underline{32.8} & \underline{54.7} & \underline{56.9} & 55.5 & \underline{77.1} & \underline{62.0} & \underline{65.5} & \underline{56.7} & \underline{56.5} & \underline{78.2} & 73.1 & 75.9 & \underline{60.4}\\
\textbf{-w/ Overall} & 37.3 & 28.8 & 51.1 & 51.9 & \underline{57.6} & 74.8 & 61.4 & 63.6 & 52.7 & 54.0 & 72.5 & \underline{73.5} & \underline{76.1} & 58.1\\

\bottomrule
\end{tabular}
\caption{\label{tab:src_gran}Performance (\%) of our framework based on GPT-3.5 with different source task granularities.}
\end{table*}

\begin{table*}[!t]
\small
\centering
\setlength{\tabcolsep}{3.3pt}
\begin{tabular}{lcccccccccccccc}
\toprule
\multirow{2}{*}{\textbf{Method}} & \multicolumn{4}{c}{\textbf{Event Relation Understand}}& \multicolumn{4}{c}{\textbf{Textual Reasoning}}& \multicolumn{3}{c}{\textbf{Sentiment Analysis}}& \multicolumn{2}{c}{\textbf{DSR}} & \multirow{2}{*}{\textit{\textbf{Avg.}}}\\
\cmidrule(lr){2-5} \cmidrule(lr){6-9} \cmidrule(lr){10-12} \cmidrule(lr){13-14}
& \textbf{Time} & \textbf{Causal} & \textbf{Coref} & \textbf{Sub-E} & \textbf{Induc} & \textbf{Deduc} & \textbf{Conj}&\textbf{SNLI}&\textbf{Twitter}&\textbf{Stance}&\textbf{Lap}&\textbf{Age}&\textbf{Dis} &  \\
\midrule
\textbf{Ours} & \textbf{43.4} & \textbf{33.0} & \textbf{57.6} & \underline{60.7} & \textbf{57.9} & \textbf{81.2} & \textbf{63.7} & \textbf{70.4} & \textbf{60.8} & \textbf{60.8} & \textbf{80.8} & \underline{76.2} & \textbf{82.4} & \textbf{63.8}\\
\textbf{-w/o Proc} & \underline{40.5} & 29.5 & \underline{57.2} & \textbf{71.7} & 54.5 & 74.3 & \underline{61.8} & \underline{69.0} & \underline{55.9} & 55.9 & 74.1 & 74.7 & 76.6 & \underline{61.2}\\
\textbf{-w/o Func} & 39.7 & \underline{30.6} & 51.0 & 59.9 & \underline{54.7} & \underline{75.9} & 60.0 & 66.1 & 55.0 & \underline{57.5} & \underline{77.3} & \textbf{77.3} & \underline{82.2} & 60.6\\
\bottomrule
\end{tabular}
\caption{\label{tab:two-similarity}Performance (\%) of our framework based on GPT-3.5 that selects source task with/without the task function and process similarity.}
\end{table*}

\begin{figure}[t]
  \centering
\includegraphics[width=1\linewidth]{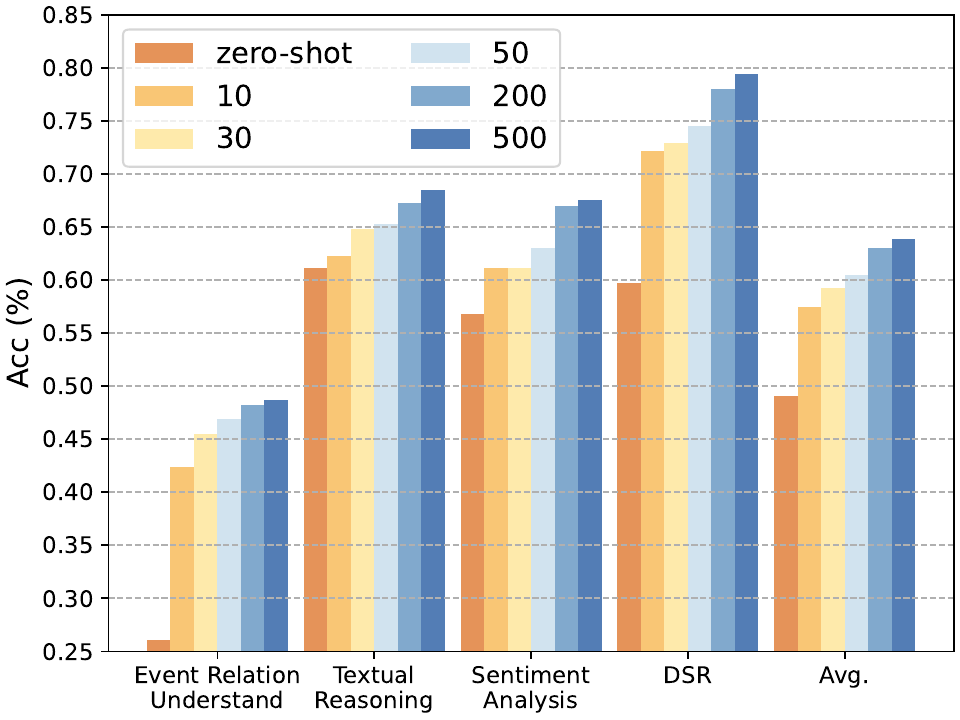}
  \caption{The impact of the number of examples in each dataset used for experience accumulation of our framework based on GPT-3.5.}
  \label{fig:datanum}
\end{figure}

\paragraph{Effect of Labeled Dataset Utilization Rate.}
\label{effect:adu}

As shown in Figure~\ref{fig:datanum}, we analyze the impact of the number of examples in each dataset used for experience accumulation.
We find that as the number of examples increases, our performance continues to improve. This is because the accumulated experience can cover more situations of the task. Furthermore, \textbf{our method remains effective even with one-tenth of the source task data.}
This indicates that while more source task data is beneficial for our framework, large-scale annotation of source task data is not essential, especially given the abundance of publicly available NLP datasets today.

\begin{figure}[t]
  \centering
\includegraphics[width=1\linewidth]{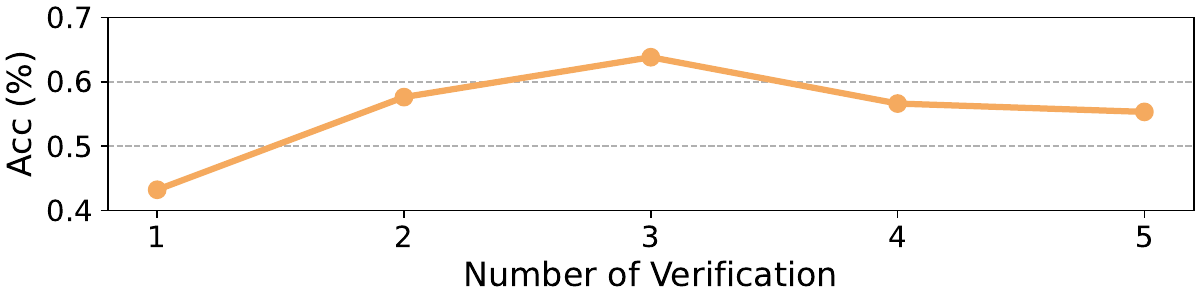}
  \caption{The impact of verification times for summarized experiences in our framework based on GPT-3.5.}
  \label{fig:check_num}
\end{figure}

\paragraph{Effect of the Number of Verification for Summarized Experiences.}
\label{effect: num-check}

As shown in Figure~\ref{fig:check_num}, we analyze the impact of the number of times the summarized experiences are validated by Prompt~\ref{pt:p4} in \S\ref{sec:eap}.
We observe that as the number decreases from 3 to 1, the performance gradually declines. This is because the summarized experience might pass 1 or 2 validations due to randomness, but its intrinsic quality is not good. Besides, as the number increases from 3 to 5, the performance also gradually decreases. This is because fewer experiences are available for transfer after validation.

\subsection{Analysis of the Source Task Selection}

\paragraph{Effect of Task Function and Process Similarity.}

As shown in Table~\ref{tab:two-similarity}, we analyze the impact of task function and process similarities on our source task selection. \textbf{-w/o Func} and \textbf{-w/o Proc} represent the exclusion of task function and process, respectively, from our source task selection. Specifically, in \S\ref{sec:sts}, only one of them is utilized for the BM25 ranking and Prompt~\ref{pt:p5}. We find that removing either one of them reduces the performance of our framework. This may be because relying solely on task functions or processes to select source tasks is not accurate enough, and a single dimension is insufficient to define the type of experience required to solve a given task. We also discuss with other retrieval-augmented methods in Appendix~\ref{app:rag}.

\begin{table*}[!t]
\small
\centering
\setlength{\tabcolsep}{2.8pt}
\begin{tabular}{lcccccccccccccc}
\toprule
\multirow{2}{*}{\textbf{Method}} & \multicolumn{4}{c}{\textbf{Event Relation Understand}}& \multicolumn{4}{c}{\textbf{Textual Reasoning}}& \multicolumn{3}{c}{\textbf{Sentiment Analysis}}& \multicolumn{2}{c}{\textbf{DSR}} & \multirow{2}{*}{\textit{\textbf{Avg.}}}\\
\cmidrule(lr){2-5} \cmidrule(lr){6-9} \cmidrule(lr){10-12} \cmidrule(lr){13-14}
& \textbf{Time} & \textbf{Causal} & \textbf{Coref} & \textbf{Sub-E} & \textbf{Induc} & \textbf{Deduc} & \textbf{Conj}&\textbf{SNLI}&\textbf{Twitter}&\textbf{Stance}&\textbf{Lap}&\textbf{Age}&\textbf{Dis} & \\
\midrule
\textbf{Minimal} & \textbf{37.0} & \underline{25.6} & \textbf{36.6} & \textbf{43.2} & \textbf{53.4} & \textbf{72.9} & \textbf{60.1} & \textbf{64.0} & \textbf{56.3} & \textbf{55.1} & \textbf{75.3} & \textbf{69.8} & \textbf{73.1} & \textbf{55.5}\\
\textbf{-w/o NoSrc} & 35.9 & 24.5 & \underline{35.5} & \underline{42.1} & 52.3 & \underline{71.8} & \underline{59.0} & \underline{62.9} & 55.2 & 54.0 & 74.2 & \underline{68.7} & \underline{72.0} & \underline{54.5}\\
\textbf{-w/o NonHard} & \underline{35.9} & \textbf{26.9} & 35.1 & 40.8 & \underline{52.6} & 70.9 & 56.4 & 62.0 & \underline{55.5} & \underline{54.3} & \underline{74.3} & 65.5 & 70.8 & 53.9\\
\bottomrule
\end{tabular}
\caption{\label{tab:skipornot}Performance (\%) of our framework based on GPT-3.5 with/without experience transfer skipping.}
\end{table*}

\begin{figure}[t]
  \centering
\includegraphics[width=1\linewidth]{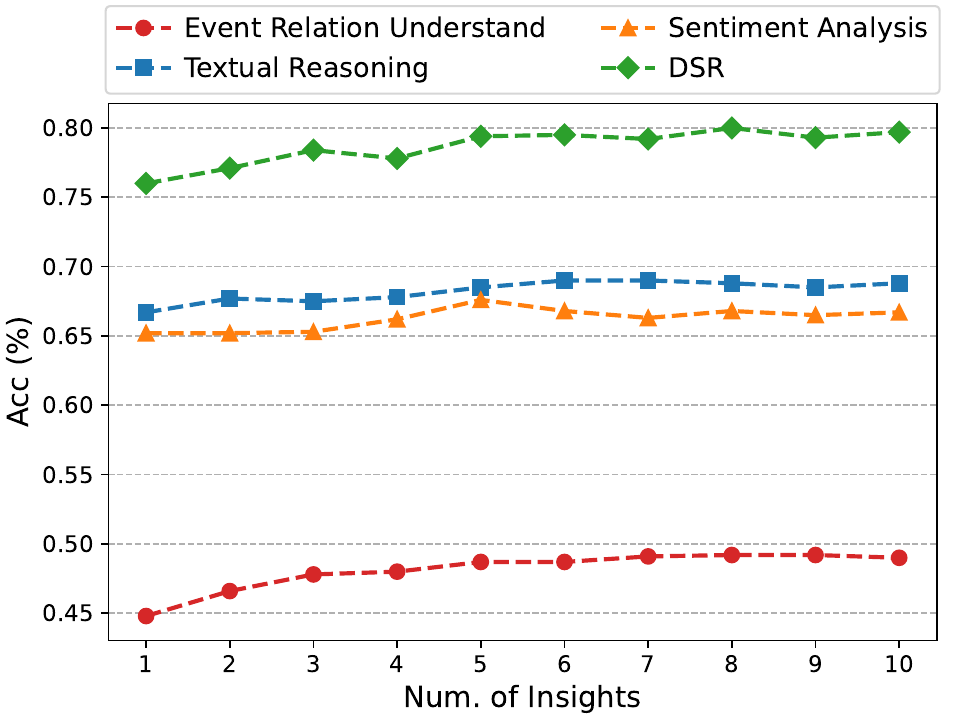}
  \caption{The impact of the number of insights used when our GPT-3.5-based framework responds to users.}
  \label{fig:M_of_finalreasoning}
\end{figure}

\begin{figure}[t]
  \centering
\includegraphics[width=1\linewidth]{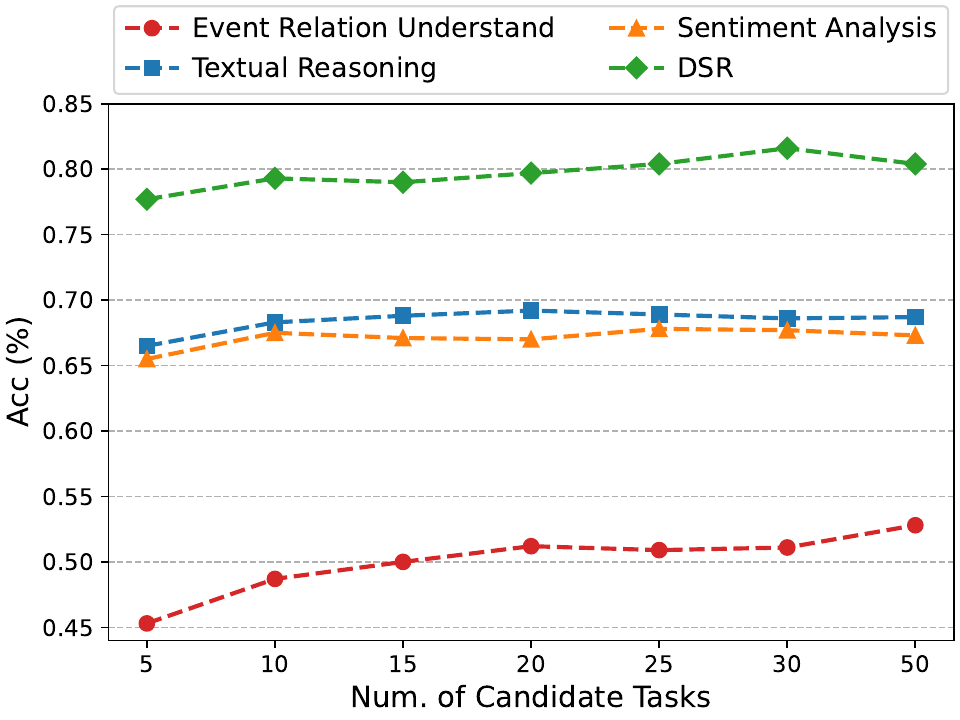}
  \caption{The effect of the number of candidates in source task selection for our GPT-3.5-based framework.}
  \label{fig:num_candidates}
\end{figure}

\paragraph{Effect of Skip Transfer.}
As shown in Table~\ref{tab:skipornot}, we analyze our two types of transfer skipping stated in \S\ref{sec:skiptrans}: 1) \textbf{-w/o NoSrc}, our framework must find at least one source task. If Prompt~\ref{pt:p5} in \S\ref{sec:sts} returns an empty list, we randomly select one from the candidates as the source task; 2) \textbf{-w/o NonHard}, our framework must select source tasks summarized from hard examples, i.e., our memory does not store non-hard examples in \S\ref{sec:eap}.
We perform experiments under the \textbf{Minimal} setting in \S\ref{effect:impact_translity}.

We find that: 1) \textbf{-w/o NoSrc} reduces model performance. This is because forcibly transferring experience when no suitable source tasks are found introduces misleading experience. 2) \textbf{-w/o NonHard} also leads to lower model performance. The reason may be that LLMs do not need additional experience to handle simple examples, and forcibly providing such experience can interfere with their reasoning. Moreover, the experience used to solve hard examples may not be suitable for simpler ones. Our framework stores non-hard examples to dynamically skip unnecessary transfer. The dynamic selective transfer capability is crucial for our framework to effectively enhance model performance.

\subsection{Analysis of Reasoning with Experience}
\label{sec:num_M}

As shown in Figure~\ref{fig:M_of_finalreasoning}, we analyze the impact of the number of insights utilized in the final reasoning Prompt~\ref{pt:p7} of our framework. It can be observed that as the number of insights increases, our performance initially improves and then stabilizes. The main reason is when there are too few insights, the transfer is insufficient, while having too many insights makes it difficult for the LLMs to effectively utilize each insight. Besides, the maximum number of insights that the model can effectively understand in a single inference may be limited.

\subsection{Effect of the Number of Candidates for Source Task Selection}
\label{app:candi_num}
As shown in Figure~\ref{fig:num_candidates}, we analyze the impact of the number  (from 5 to 50) of candidates in source task selection for our framework based on GPT-3.5.

We find that as the number increases, our performance improves slightly. This may be due to the additional potential effective source task experiences provided. However, it also proportionally increases the inference time, which is critical for user satisfaction. Therefore, in our framework, we set the number of candidates to 10 as a tradeoff.

\section{Related Work}

\subsection{Experiential Learning for NLP Tasks}
\label{related:EL}
Training or fine-tuning LLMs like ChatGPT can be quite challenging \citep{openai2024gpt4technicalreport}.
To enhance LLMs during their inference phase, previous research has provided LLMs with textual task-solving experiences through prompts.

Early research primarily relied on manually crafting experiences \citep{wei2022chain,kong2023better}, while recent studies have shifted towards leveraging LLMs to automatically summarize experiences from existing labeled NLP datasets for each task \citep{zhao2023expel,du-etal-2023-task,yang-etal-2023-failures,chen2024grimoire,tong-etal-2024-llms}.
Furthermore, for each user query, SEGPT \citep{gao-etal-2024-self-evolving} first uses LLMs to generate a small pseudo-dataset from scratch (taking 3 to 8 minutes), and then summarizes experience from the dataset.

However, these methods require substantial costs to gain experience for just a single task. Recently, as the types of new tasks in user queries continue to increase, these methods are becoming impractical.
In contrast, our framework transfers existing task experiences to new tasks, acquiring new experience without substantial human labor and time costs.

Additionally, SEGPT \citep{gao-etal-2024-self-evolving} also designed an auxiliary module to rewrite shared common insights across multiple tasks for the user query, which helps in the generation of pseudo-datasets. This can be seen as an oversimplification of experience transfer.
However, these common insights are often too trivial. Their experiments show that relying solely on such common insights cannot significantly improve the performance of LLMs.
Furthermore, for the user query of new tasks, \citet{chatterjee-etal-2024-language} require users to manually provide task-level instructions and multiple similar queries. They then leverage existing datasets from other tasks for cross-task in-context learning (ICL) to label these queries, which are then used as ICL demonstrations to infer the user query.
This method relies on examples rather than experience, making it highly sensitive to format differences across tasks, and it also relies on extra manual labor.

As far as we know, our work is the first to systematically explore the textual experience transfer ability of LLMs in NLP.

\subsection{Experiential Agents Guided by Interactive Environments}
Beyond NLP, researchers in other fields are exploring how to guide LLM-based agents to autonomously summarize experiences from interactive environments.

\citet{chen2023introspective} guided an agent to learn cooking processes through a cooking simulation game.
\citet{wang2023voyager} and \citet{zhu2023ghost} developed agents in the game ``Minecraft'', enabling them to learn how to accomplish various in-game objectives, such as building a house.
\citet{wen2023dilu} and \citet{fu2024drive} utilized a simulated driving environment to allow LLMs to summarize driving experiences.
\citet{ma2024largelanguagemodelsplay} designed agents to learn how to play the game ``StarCraft II''.
 
However, these methods rely on interactive environments that are not accessible for most NLP tasks. In contrast, our approach enables the transfer of experience across NLP tasks, acquiring new task experience without relying on such environments.

\section{Conclusion}

In this work, we propose an autonomous experience transfer framework based on LLMs. It automatically summarizes experiences from existing labeled datasets and transfers them to new tasks for high-quality responses. Considering the continually increasing diversity of task types in user queries to LLMs, our framework effectively reduces the human labor and time costs required by previous methods. Experiments demonstrate that our framework can reliably perform experience transfer to improve the performance of LLMs.

\section*{Acknowledgments}

The research in this article is supported by the New Generation Artificial Intelligence of China (2024YFE0203700), National Natural Science Foundation of China under Grants U22B2059 and
62176079.

\section*{Limitations}
Same as previous experiential learning studies, our work mainly focuses on classification tasks. We provide further discussion on how to extend our framework to other task types in Appendix~\ref{app:othertasktype}.

Besides, our framework utilizes as much data as possible during the source experience accumulation phase, though this may not be economically optimal. A dynamic control of the data usage for each dataset could help achieve an optimal balance between costs and performance.

In addition, our experiment shows that for tasks that LLMs have already mastered proficiently, the performance improvement brought by additional experience may not be significant.

\bibliography{custom}

\clearpage

\appendix
\section*{Appendix}
\tableofcontents

\clearpage

\begin{table*}[!t]
\small
\centering
\setlength{\tabcolsep}{3.3pt}
\begin{tabular}{lcccccccccccccc}
\toprule
\multirow{2}{*}{\textbf{Method}} & \multicolumn{4}{c}{\textbf{Event Relation Understand}}& \multicolumn{4}{c}{\textbf{Textual Reasoning}}& \multicolumn{3}{c}{\textbf{Sentiment Analysis}}& \multicolumn{2}{c}{\textbf{DSR}} & \multirow{2}{*}{\textit{\textbf{Avg.}}}\\
\cmidrule(lr){2-5} \cmidrule(lr){6-9} \cmidrule(lr){10-12} \cmidrule(lr){13-14}
& \textbf{Time} & \textbf{Causal} & \textbf{Coref} & \textbf{Sub-E} & \textbf{Induc} & \textbf{Deduc} & \textbf{Conj}&\textbf{SNLI}&\textbf{Twitter}&\textbf{Stance}&\textbf{Lap}&\textbf{Age}&\textbf{Dis} &  \\
\midrule
\textbf{Zero-shot} & 37.3 & 24.8 & 11.4 & 30.9 & 52.4 & 70.3 & 59.6 & 62.0 & 49.3 & 50.0 & 71.2 & 58.0 & 61.4 & 49.1\\
\textbf{ZS-CoT} & 38.0 & 24.4 & 30.8 & 43.3 & 53.6 & 76.2 & 55.6 & 63.6 & 53.3 & 56.3 & 74.4 & 69.6 & 74.5 & 54.9\\
\midrule
\textbf{Normal} & \textbf{43.4} & \underline{33.0} & \underline{57.6} & \textbf{60.7} & \textbf{57.9} & \textbf{81.2} & \textbf{63.7} & \textbf{70.4} & \textbf{60.8} & \textbf{60.8} & \textbf{80.8} & \underline{76.2} & \underline{82.4} & \underline{63.8}\\
\textbf{Low} & \underline{40.3} & 30.9 & 44.7 & 54.8 & \underline{57.8} & \underline{79.3} & \underline{62.5} & \underline{68.7} & \underline{60.7} & 58.2 & \underline{78.5} & 73.1 & 76.2 & 60.4\\
\textbf{Minimal} & 37.0 & 25.6 & 36.6 & 43.2 & 53.4 & 72.9 & 60.1 & 64.0 & 56.3 & 55.1 & 75.3 & 69.8 & 73.1 & 55.5\\
\midrule
\textbf{SameSet} & 39.3 & \textbf{37.6} & \textbf{70.1} & \underline{59.8} & 56.9 & 76.4 & 59.9 & 67.7 & 59.0 & \underline{58.9} & 77.6 & \textbf{84.4} & \textbf{82.7} & \textbf{63.9}\\
\bottomrule
\end{tabular}
\caption{\label{tab:intrataskexperience}Performance (\%) of our framework based on GPT-3.5 with experience learned from the same dataset or transferred from other datasets.}
\end{table*}

\section{Comparison of Experience Gained from Same-Dataset and Cross-Dataset Transfer}
\label{app:samesetexp}

As shown in Table~\ref{tab:intrataskexperience}, we analyze the performance differences of our framework when using experiences learned from the same dataset and experiences transferred across different datasets. The \textbf{Normal}, \textbf{Low} and \textbf{Minimal} settings are same as in \S\ref{effect:impact_translity}. And the \textbf{SameSet} means that each test example is only allowed to select source tasks from the same dataset as itself, i.e., experience reuse within the same dataset. We can find that:

Firstly, the performance is similar under both the \textbf{Normal} setting and the \textbf{SameSet} setting. This is because our framework enables reliable experience transfer, and when appropriate source tasks are available, it can also obtain effective experience for the target task.

Besides, it can be observed that our framework performs better in the \textbf{Normal} setting than in the \textbf{SameSet} setting on some datasets. This may be because difficult examples are more specific and occur less frequently in a given dataset, making it hard to find identical cases within the same dataset to draw on for solutions. However, in a cross-task setting, although the source tasks may differ from the current target task, there are multiple source tasks, which allows the transferred experience to offer greater diversity than what could be retrieved from a single dataset. This enables the model to approach the current problem from more angles, potentially leading to better solutions.

Furthermore, the performance in \textbf{Low} and \textbf{Minimal} settings is not as good as in \textbf{SameSet}. This is because the effectiveness of experience transfer is limited by whether suitable source tasks exist. For a new task that is dissimilar to any previous tasks, experience transfer may fail.

\begin{figure}[t]
  \centering
\includegraphics[width=1\linewidth]{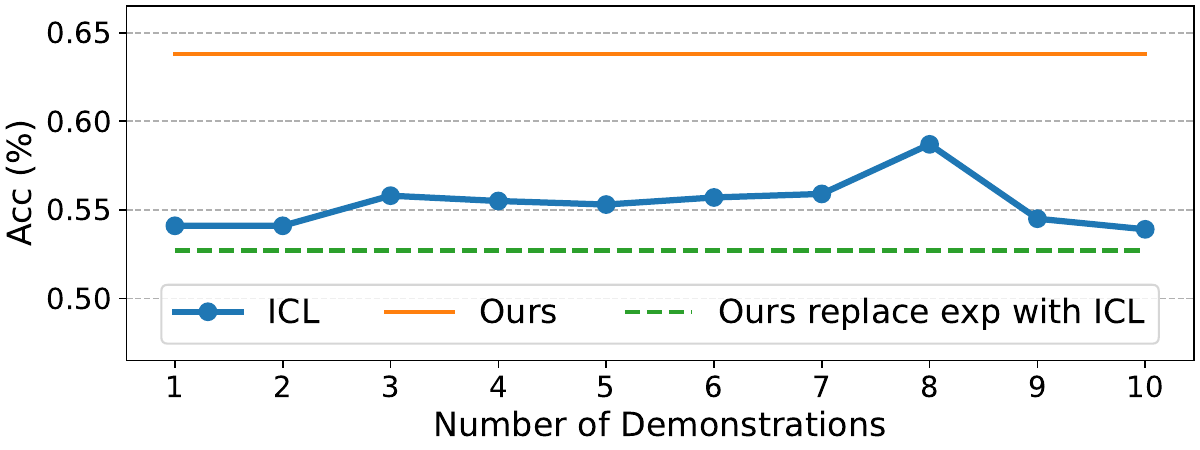}
  \caption{Performance of the ICL and out framework with/without experience based on GPT-3.5.}
  \label{fig:fewshot}
\end{figure}

\section{Experience vs. Demonstrations}
\label{evse}
In this section, we discuss which way of leveraging existing data for our framework is better: 1) providing demonstrations to LLMs through ICL; 2) leveraging the textual experiences we focus on. As shown in Figure~\ref{fig:fewshot}, we compare three setups: 1) the standard \textbf{ICL}, where demonstrations are concatenated before the current query; 2) \textbf{Ours}: our framework ExpeTrans; 3) \textbf{Ours replace exp with ICL}: in \S\ref{sec:rwe}, we do not use experiences for reasoning but directly utilize selected source task examples as demonstrations to perform ICL inference. Note that the horizontal axis of Figure~\ref{fig:fewshot} is unrelated to the latter two setups.

It can be observed that using abstract experiences rather than concrete demonstrations is better for enhancing model performance. This is because, compared to demonstrations, experiences are more generalizable, and subtle differences in format and input-output requirements in demonstrations can interfere with the reasoning. Additionally, \citet{chen2024grimoire} have also demonstrated that such textual experiences lead to better improvements for LLMs compared to demonstrations.

\section{Baseline: Cross-ICL}
\label{app:cross-icl}

\citet{chatterjee-etal-2024-language} tried to achieve cross-task transfer through cross-task ICL demonstrations. So, is it unnecessary to rely on experience for transfer, and can we simply depend on demonstrations? To explore this question, we designed Cross-ICL as a baseline method.

Specifically, Cross-ICL performs the ICL in the cross-task setting.
In our experiments, for each query, the top 5 examples with the highest semantic similarity to that query are selected from the other 12 datasets to serve as demonstrations for ICL. Specifically, we obtain the semantic vectors of the examples and query using Sentence-BERT and calculate the cosine similarity of the semantic vectors as the semantic similarity between them.
Besides, for a fair comparison with our framework and other baseline methods, we use ChatGPT to generate thought processes for each demonstrations, thereby performing cross-task ICL with CoT. Otherwise, the performance of Cross-ICL will significantly decrease, leading to unfair comparisons.

\begin{table*}[!t]
\small
\centering
\setlength{\tabcolsep}{3pt}
\begin{tabular}{lcccccccccccccc}
\toprule
\multirow{2}{*}{\textbf{Method}} & \multicolumn{4}{c}{\textbf{Event Relation Understand}}& \multicolumn{4}{c}{\textbf{Textual Reasoning}}& \multicolumn{3}{c}{\textbf{Sentiment Analysis}}& \multicolumn{2}{c}{\textbf{DSR}} & \multirow{2}{*}{\textit{\textbf{Avg.}}}\\
\cmidrule(lr){2-5} \cmidrule(lr){6-9} \cmidrule(lr){10-12} \cmidrule(lr){13-14}
& \textbf{Time} & \textbf{Causal} & \textbf{Coref} & \textbf{Sub-E} & \textbf{Induc} & \textbf{Deduc} & \textbf{Conj}&\textbf{SNLI}&\textbf{Twitter}&\textbf{Stance}&\textbf{Lap}&\textbf{Age}&\textbf{Dis} &  \\
\midrule
\textbf{Ours + BEI} & \textbf{46.5} & \textbf{35.8} & \textbf{61.1} & \textbf{65.4} & \textbf{60.5} & \textbf{83.1} & \textbf{65.5} & 70.1 & \underline{63.2} & \underline{62.4} & \textbf{82.0} & \underline{79.1} & \textbf{84.0} & \textbf{66.1}\\
\textbf{Ours} & \underline{43.4} & \underline{33.0} & 57.6 & 60.7 & 57.9 & \underline{81.2} & 63.7 & \underline{70.4} & 60.8 & 60.8 & \underline{80.8} & 76.2 & \underline{82.4} & \underline{63.8}\\
\textbf{BEI} & 41.1 & 30.6 & \underline{59.8} & \underline{63.1} & \underline{58.8} & 70.5 & 60.1 & 70.3 & 58.0 & 56.1 & 77.1 & \textbf{80.6} & 81.6 & 62.1\\
\textbf{REFEED} & 42.8 & 30.4 & 44.2 & 49.4 & 57.4 & 72.9 & 64.0 & 67.8 & \textbf{63.3} & \textbf{62.8} & 79.4 & 71.5 & 68.4 & 59.6\\
\textbf{DPR} & 40.5 & 28.9 & 52.6 & 56.3 & 57.7 & 75.5 & \underline{64.1} & \textbf{71.9} & 61.2 & 60.5 & 78.7 & 72.4 & 71.4 & 60.9\\
\bottomrule
\end{tabular}
\caption{\label{tab:rag}Performance (\%) of our framework based on GPT-3.5 compared to other retrieval variants.}
\end{table*}

\begin{table*}[!t]
\small
\centering
\setlength{\tabcolsep}{3.25pt}
\begin{tabular}{lcccccccccccccc}
\toprule
\multirow{2}{*}{\textbf{Method}} & \multicolumn{4}{c}{\textbf{Event Relation Understand}}& \multicolumn{4}{c}{\textbf{Textual Reasoning}}& \multicolumn{3}{c}{\textbf{Sentiment Analysis}}& \multicolumn{2}{c}{\textbf{DSR}} & \multirow{2}{*}{\textit{\textbf{Avg.}}}\\
\cmidrule(lr){2-5} \cmidrule(lr){6-9} \cmidrule(lr){10-12} \cmidrule(lr){13-14}
& \textbf{Time} & \textbf{Causal} & \textbf{Coref} & \textbf{Sub-E} & \textbf{Induc} & \textbf{Deduc} & \textbf{Conj}&\textbf{SNLI}&\textbf{Twitter}&\textbf{Stance}&\textbf{Lap}&\textbf{Age}&\textbf{Dis} &  \\
\midrule
\rowcolor{gray!20} 
\multicolumn{15}{c}{\textbf{\emph{Experiments based on \texttt{gemma-1.1-7b-it}}}}\\
\textbf{Zero-shot} & 33.2 & 19.0 & 7.8 & \textbf{60.2} & 54.4 & 55.4 & 35.0 & 36.5 & 57.2 & \underline{47.8} & 72.0 & 52.1 & 46.7 & 44.4\\
\textbf{ZS-CoT} & \textbf{37.3} & \underline{21.9} & \underline{10.4} & 42.2 & \textbf{60.2} & \textbf{64.3} & \textbf{55.7} & \textbf{51.3} & \textbf{60.1} & 45.6 & \textbf{73.0} & \underline{59.0} & \underline{56.0} & \underline{49.0}\\
\textbf{Ours} & \underline{35.6} & \textbf{25.2} & \textbf{20.0} & \underline{44.1} & \underline{59.6} & \underline{56.4} & \underline{50.7} & \underline{48.4} & \underline{58.5} & \textbf{49.0} & \underline{72.9} & \textbf{64.3} & \textbf{63.5} & \textbf{49.9}\\
\midrule
\rowcolor{gray!20} 
\multicolumn{15}{c}{\textbf{\emph{Experiments based on \texttt{chatglm3-6b}}}}\\
\textbf{Zero-shot} & 22.6 & 14.0 & \textbf{28.7} & \textbf{27.9} & \textbf{55.3} & 51.3 & \textbf{56.9} & \textbf{75.9} & 54.5 & 46.8 & 71.4 & 60.4 & \textbf{63.3} & \textbf{48.4}\\
\textbf{ZS-CoT} & \textbf{32.8} & \textbf{29.6} & 3.5 & 17.0 & 54.4 & \textbf{55.4} & 43.7 & 48.3 & \underline{58.2} & \textbf{49.1} & \underline{73.5} & \textbf{63.1} & \underline{61.0} & 45.4\\
\textbf{Ours} & \underline{29.1} & \underline{27.3} & \underline{22.7} & \underline{21.5} & \underline{54.6} & \underline{54.1} & \underline{44.7} & \underline{49.0} & \textbf{58.7} & \underline{49.0} & \textbf{75.8} & \underline{61.1} & 57.5 & \underline{46.5}\\
\midrule
\rowcolor{gray!20} 
\multicolumn{15}{c}{\textbf{\emph{Experiments based on \texttt{deepseek-llm-7b-chat}}}}\\
\textbf{Zero-shot} & 23.6 & 12.6 & 16.1 & 38.3 & 54.0 & 48.8 & 49.9 & 61.5 & \textbf{62.6} & \textbf{49.0} & 73.6 & 57.4 & 57.0 & 46.5\\
\textbf{ZS-CoT} & \underline{27.9} & \underline{23.2} & \underline{19.8} & \textbf{52.8} & \underline{56.8} & \underline{56.4} & \underline{60.5} & \underline{74.7} & 60.9 & 43.9 & \textbf{76.5} & \underline{63.8} & \underline{61.4} & \underline{52.2}\\
\textbf{Ours} & \textbf{28.8} & \textbf{25.1} & \textbf{20.0} & \underline{50.5} & \textbf{58.8} & \textbf{59.4} & \textbf{62.4} & \textbf{77.2} & \underline{61.8} & \underline{47.2} & \underline{76.1} & \textbf{64.9} & \textbf{62.3} & \textbf{53.4}\\
\midrule
\rowcolor{gray!20} 
\multicolumn{15}{c}{\textbf{\emph{Experiments based on \texttt{llama-3.1-8B-instruct}}}}\\
\textbf{Zero-shot} & \textbf{43.6} & \textbf{41.6} & \underline{58.1} & 43.1 & \textbf{60.5} & \textbf{76.2} & 43.6 & 54.4 & 47.3 & 48.4 & 71.5 & \underline{65.9} & 61.3 & 55.0\\
\textbf{ZS-CoT} & 36.8 & 30.4 & 55.3 & \underline{48.2} & 57.5 & 75.6 & \underline{57.2} & \underline{61.0} & \underline{60.4} & \underline{58.5} & \underline{79.5} & 62.0 & \underline{64.2} & \underline{57.4}\\
\textbf{Ours} & \underline{38.8} & \underline{35.0} & \textbf{65.0} & \textbf{58.5} & \underline{60.0} & \underline{76.0} & \textbf{58.9} & \textbf{61.0} & \textbf{60.9} & \textbf{59.5} & \textbf{80.5} & \textbf{70.3} & \textbf{75.0} & \textbf{61.5}\\
\midrule
\rowcolor{gray!20} 
\multicolumn{15}{c}{\textbf{\emph{Experiments based on \texttt{DeepSeek V3}}}}\\
\textbf{Zero-shot} & 42.9 & 44.2 & 69.4 & \underline{100.0} & \underline{77.2} & \textbf{85.3} & \underline{69.6} & 54.7 & \textbf{67.8} & \textbf{68.7} & 84.9 & \underline{85.4} & \underline{94.0} & 72.6\\
\textbf{ZS-CoT} & \underline{45.1} & \underline{47.3} & \textbf{84.9} & \textbf{100.0} & \textbf{79.0} & \underline{82.5} & \textbf{75.5} & \textbf{55.7} & 66.4 & \underline{68.4} & \underline{86.0} & 76.7 & 91.2 & \underline{73.7}\\
\textbf{Ours} & \textbf{46.6} & \textbf{59.2} & \underline{80.6} & 98.5 & 76.4 & 81.8 & 66.2 & \underline{55.6} & \underline{67.7} & 68.1 & \textbf{87.7} & \textbf{88.7} & \textbf{94.5} & \textbf{74.7}\\

\midrule

\rowcolor{gray!20} 
\multicolumn{15}{c}{\textbf{\emph{Experiments based on \texttt{GPT-4}}}}\\
\textbf{Zero-shot} & 39.5 & 37.9 & 86.5 & 63.2 & 50.7 & 97.2 & 69.5 & 78.2 & 60.2 & 61.9 & 79.7 & 88.5 & 94.9 & 69.8\\
\textbf{ZS-CoT} & 40.8 & 35.5 & 87.7 & \underline{68.8} & 51.5 & \textbf{98.5} & 71.0 & 78.5 & 61.3 & 65.9 & 81.0 & 86.2 & 94.3 & \underline{70.8}\\
\textbf{Self-EXP} & 35.9 & 36.7 & 87.5 & 65.1 & 48.2 & 97.0 & 65.8 & 79.4 & \underline{62.8} & 63.3 & \underline{81.1} & 89.9 & 93.9 & 69.7\\
\textbf{SEGPT-T} & 41.2 & 32.7 & \underline{88.1} & 65.6 & 50.1 & 96.8 & \underline{71.5} & 77.4 & 60.0 & \underline{67.4} & 80.6 & 86.3 & 94.5 & 70.2\\
\textbf{Cross-ICL} & 35.9 & 40.5 & 87.2 & 51.1 & \underline{51.5} & 95.5 & 69.8 & \underline{81.9} & 62.0 & 66.2 & 79.8 & 94.0 & \textbf{97.4} & 70.1\\
\textbf{Cross-ICL*} & \underline{41.6} & \textbf{43.3} & 79.3 & 49.8 & 50.3 & 95.7 & 68.3 & 77.9 & 60.7 & 62.1 & 78.2 & \underline{94.7} & \underline{97.1} & 69.2\\
\textbf{Ours} & \textbf{44.5} & \underline{42.7} & \textbf{91.4} & \textbf{69.2} & \textbf{56.6} & \underline{98.3} & \textbf{75.8} & \textbf{83.3} & \textbf{66.7} & \textbf{70.9} & \textbf{84.4} & \textbf{95.8} & 96.3 & \textbf{75.1}\\
\bottomrule
\end{tabular}
\caption{\label{tab:smallllms}Performance (\%) of our framework based on other LLMs.}
\end{table*}

\section{Discussion with Other Retrieval-Augmented Methods}
\label{app:rag}
Our framework covers various aspects of experience learning and consists of: experience acquisition and experience application, with similar task retrieval being one of the three sub-modules of the latter. Given that task retrieval is only a part of half of our framework, we did not employ the retrieval-augmented generation (RAG) method as a baseline for the entire framework.
However, they can be baselines for our task retrieval as the ablation study.
Specifically, we review the RAG suvery \citep{fan2024survey} and replace our task retrieval  with other baseline methods in the RAG domain: 1) \textbf{DPR} \citep{karpukhin2020dense}, a widely-used dense retriever; 2) \textbf{REFEED} \citep{yu2023improving}, a pre-retrieval enhancement method that leverages the initial responses of LLMs to improve retrieval; 3) \textbf{BGE-EN-ICL (BEI)} \citep{li2024making}, one of the most powerful dense retrievers to date. By providing retrieval examples, it can effectively enhance retrieval; 4) \textbf{Ours + BEI}: we replaced the BM25 in our task retrieval module with BGE-EN-ICL. Results are shown in Table~\ref{tab:rag}.
It can be observed that, our task retrieval module, which is specifically designed for experience transfer, achieves the best average performance. Besides, further improvements are achieved by Ours + BEI. The main reason is the stronger capability of BEI compared to the BM25 we used in our framework.

\section{Performance on Other LLMs}
\label{app:smallllm}
In this section, we discuss the performance of our framework based on other smaller or larger LLMs. 

Typically, smaller LLMs are used for fine-tuning rather than experiential learning.
Our work focuses on experiential learning. This line of research investigates black-box API-based LLMs such as ChatGPT, which are often difficult to fine-tune and therefore particularly benefit from experiential learning. Therefore, previous studies do not compare with fine-tuning methods, and our paper follows the same setup.
Besides, comparing to fine-tuning methods might be inherently unfair. Fine-tuning methods can directly update model parameters, leading to a more substantial impact. However, fine-tuning often suffers from catastrophic forgetting, especially critical issue for large-scale LLMs like ChatGPT. In contrast, experiential learning does not update model parameters, preserving the generalization ability.

To provide more insights into experience transfer, we present the performance of our framework on these LLMs: 1) \textbf{\texttt{gemma-1.1-7b-it}} \citep{gemmateam2024gemmaopenmodelsbased}; 2) \textbf{\texttt{chatglm3-6b}} \citep{glm2024chatglmfamilylargelanguage}; 3) \textbf{\texttt{deepseek-llm-7b-chat}} \citep{deepseekai2024deepseekllmscalingopensource}; 4) \textbf{\texttt{llama-3.1-8B-instruct}} \citep{dubey2024llama}; 5) \textbf{\texttt{DeepSeek-V3}} \citep{liu2024deepseek}; 6) \textbf{\texttt{GPT-4}} \citep{hurst2024gpt}. We also perform a brief test with a slow think LLM.
Results are shown in Table~\ref{tab:smallllms}. It can be observed that:

Firstly, the model does not necessarily perform better with our method just because it is smaller. Our error analysis finds that they may perform poor self-reflection, making it hard to obtain good experiences from source data (a case for gemma is shown in Case~\ref{casestudy:case6}). Therefore, the effectiveness on smaller models depends on their ability to reflect.

Besides, Our framework based on DeepSeek V3 does not achieve satisfactory performance. One possible reason is that prompt effectiveness varies significantly across different base LLMs. Meanwhile, we also conduct a small-scale testing on the slow think LLMs (distilled 7B DeepSeek R1) using 10 samples for each task. Our observations reveal similar prompt adaptation issues, preventing effective improvement in model performance (only a 2.7\% performance gain is observed). When switching between different base LLMs families, our framework may require adjusting the prompts to align with their respective formats-particularly in controlling the parsing of output content.

Additionally, our improvement on GPT-4 is less significant compared to GPT-3.5. The reason may be that GPT-4 is more proficient at tasks, resulting in more experience already being mastered by GPT-4. 
Transferred experience may perform better on more novel and unexpected tasks for LLMs. Considering the increased costs associated with larger LLMs, the value of our method may diminish when applied to such larger LLMs. Our future work will focus on further enhancing the performance of our framework on smaller LLMs while reducing its computational cost on larger ones.

\begin{figure}[t]
  \centering
\includegraphics[width=1\linewidth]{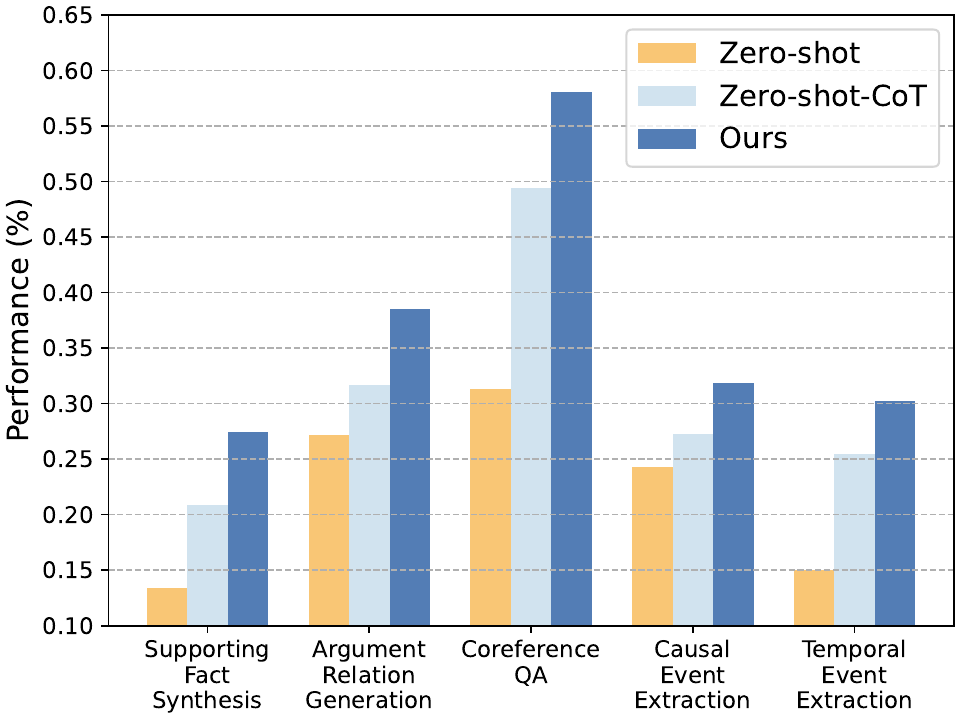}
  \caption{Performance (\%) of our framework based on GPT-3.5 in different task types. The three on the left are generative tasks, for which we report the ROUGE-L score. The two on the right are extractive tasks, for which we report the F1-score.}
  \label{fig:othertasktypes}
\end{figure}

\section{Extend to Other Task Types}
\label{app:othertasktype}
In this section, we further discuss extending our framework to more task types.
Previous experiential learning studies are evaluated on classification tasks, our work follows their setup. However, our framework is not limited to classification tasks. 
It do not impose restrictions on task output formats. There is only one exception: in our Experience Accumulation (\S\ref{sec:stea}), where the prediction quality of LLMs is evaluated based on the golden label for source task data though ``Accuracy''. To extend to other task types, our framework requires different evaluation metrics (like ``ROUGE-L'' for generative tasks or ``F1-score'' for extractive tasks).

As a demonstration, we report the results on five non-classification tasks in Figure~\ref{fig:othertasktypes}. These tasks include \textbf{generative tasks} proposed by \citet{weston2015towards}: 1) \textbf{Supporting Fact Synthesis}: Given a context, generate facts supported by the context about the target entity; 2) \textbf{Argument Relation Generation}: Given a context, generate entities that are in a required relationship with the target entity; 3)~\textbf{Coreference QA}: Given a context, answer the question depends on the understanding of coreference. And \textbf{extractive tasks} proposed by \citet{wang-etal-2022-maven}: 1) \textbf{Causal Event Extraction}: Given a text, extract multiple event pairs that exhibit causal relationships. The key challenge lies in identifying implicit causal relationships triggered by no obvious cues \citep{gao2024event}; 2) \textbf{Temporal Event Extraction}: Given a text, extract multiple event pairs that exhibit temporal relationships.
For each task, 500 examples are testing.
Our framework uses ROUGE-L and relaxed F1-score \citep{gao2024enhancing} as evaluation metrics for these tasks.
We find that by making adjustments to the evaluation metrics for model response quality, our framework has the potential to be extended to other task types.

\section{Additional Information on Responsible NLP Research}
\label{appendix:checklist}

\paragraph{Potential Risks.} In general, our work itself does not pose potential risks, but it may be used for malicious purposes. For example, applying experience transfer technology to large model attack scenarios may create risks.

\paragraph{Use Scientific Artifacts.} As shown in Section \S\ref{datasets}, we use 13 NLP benchmark datasets. They are all in English language.
They are all allowed to be used for scientific research. They are all widely used datasets in the NLP community and have been vetted by the NLP community that they do not information that names or uniquely identifies individual people or offensive content. In addition, we sampled 100 examples from each dataset for manual inspection to check that these datasets did not contain objectionable content.
The MAVEN-ERE benchmark follows GPL-3.0 license. The bAbI benchmark follows BSD license. The Conj dataset follows MIT license. The SNLI dataset follows CC BY-SA 4.0 license. The Twitter17 and the Stance16 dataset follow MIT license. The Lap14 dataset follows CC-BY-4.0 license. The BBQ-Lite dataset follows CC-BY-4.0 license. Our use of these datasets are consistent with their intended purpose. There is no significant bias in the population distribution of these datasets.

\paragraph{Parameters For Packages} The version of packages we used: openai=1.3.3, rank-bm25=0.2.2, numpy=1.24.4, sentence-transformers=3.0.1, torch=2.1.1, spacy=3.7.5. More details can be found in our supplementary materials.

\paragraph{Scale of the Models Used} Note that the exact scale of ChatGPT is unknown.
Here, based on previous analyses, we estimate that GPT-3.5 may contain 175 billion parameters.

\paragraph{AI Assistants in Writing.}
We use ChatGPT to help check for grammatical errors and provide suggestions for improving language expression.

\section{Background discussion on SMT}
Structural Mapping Theory (SMT) is a cognitive theory that explains the mechanisms of analogical reasoning, which serves as the foundation of transfer learning. In SMT, human analogical reasoning is considered to operate by aligning the relational structures between the source and target domains, emphasizing deep, interconnected relational mappings rather than superficial surface similarities. SMT also has a profound impact in fields such as psychology and education, providing a theoretical basis for understanding human analogical and transfer learning abilities. 
When aligning relational structures across different tasks, humans typically focus on key structures such as task goals and execution processes. In this paper, we primarily concentrate on these two main structure features while selectively disregarding others like task domains and languages to avoid excessive complexity.

\section{Transfer Between Tasks with Asymmetric Difficulty}

\begin{table}[!t]
\small
\centering
\begin{tabular}{lcc}
\toprule
\textbf{Method} & \textbf{Twitter17}& \textbf{GSM8K} \\
\midrule

\textbf{Zero-Shot}	& 0.493	& 0.785 \\
\textbf{Zero-shot-CoT}	& \underline{0.533}	& \underline{0.789} \\
\textbf{Ours}	& 0.\textbf{575}	& \textbf{0.815} \\
\bottomrule
\end{tabular}
\caption{\label{tab:asymme}Performance (\%) of transfer between tasks with asymmetric difficulty.}
\end{table}

In this section, we discuss transfer learning between tasks of different difficulty levels, using sentiment analysis and mathematical reasoning as examples. Typically, mathematical reasoning is considered a challenging task that requires slow, deliberate thinking to solve. Specifically, we conduct experiments using GSM8K \citep{cobbe2021training} (a commonly used math dataset) and Twitter17 (the sentiment analysis dataset in previous sections), randomly selecting 500 samples from each dataset and testing them with GPT-3.5-turbo.

As shown in Table~\ref{tab:asymme}, even though the two tasks are highly less similar, our framework still strives to transfer useful experience. It can effectively gather experience from the more complex task, i.e., mathematical reasoning task. Furthermore, it can be observed that the transfer is asymmetric, which may primarily be due to the complex logical thoughts in mathematical reasoning enhancing the model's level of detail when considering sentiment analysis problems. On the other hand, sentiment analysis involves more perception of context and human emotions, which is difficult to transfer to mathematical reasoning.

\section{Case Study}
\label{app:casestudy}

\begin{table*}[!t]
\small
\centering
\begin{tabular}{lcccc}
\toprule
\textbf{Proportion} & \textbf{Event Relation} & \textbf{Textual Reasoning} &  \textbf{Sentiment Analysis} & \textbf{DSR} \\
\midrule
\textbf{GPT-3.5} & 21.7 & 26.5 & 24.1 & 27.7 \\
\textbf{GPT-4} & 17.9 & 33.3 & 20.5 & 28.3 \\
\bottomrule
\end{tabular}
\caption{\label{tab:erro_p_distrib}Proportion (\%) of cases that incorporating experience explicitly disrupts the originally correct reasoning.}
\end{table*}

In this section, we conduct a case study on how experience influences the reasoning process of LLMs, demonstrating how experience can correct initially incorrect inference, and in rare instances, how experience might hurt the reasoning process.
For each example, we provide both the responses with or without the experience. \emph{Due to space limitations, the examples are shown starting from the page 19.}

For Case-\ref{casestudy:case1}, the correct answer is B. Zero-shot-CoT selected the wrong answer. This is because, although it engaged in step-by-step reasoning, the reasoning was rather shallow. In fact, the first two steps merely restated the input question, demonstrating insufficient depth of thought. In contrast, after incorporating experience, the model produced the correct response. It can be observed that the added experience enabled the LLMs to consider the problem from more dimensions, leading to a more detailed analytical process. \textbf{This refinement in reasoning granularity improved the performance of LLMs}. Additionally, it is worth noting that \textbf{the LLMs do not rigidly utilize all the provided experiences or copy them verbatim}. Instead, they use experience as inspiration to enrich the reasoning process.

For Case-\ref{casestudy:case2}, the correct answer is C. Zero-shot-CoT mistakenly assumes causal relationships between events, even when none actually exist. This phenomenon is known as \textbf{causal hallucination} \citep{gao-etal-2023-chatgpt}, i.e., LLMs are prone to fabricating causality between events.
According to \citet{gao-etal-2023-chatgpt}, this may be because, in the natural texts used during the pretraining of LLMs, causal relationships are often explicitly expressed with trigger words such as ``cause'' or ``lead to'', whereas non-causal relationships are typically not emphasized. This reporting bias is a potential reason of causal hallucination. 
In contrast, after incorporating experience, the model selects the correct answer. This improvement is due to the experience \textbf{providing a detailed analytical approach}, guiding the model on key aspects to focus on in causal reasoning. Additionally, the experience explicitly \textbf{reminds the model to ``exercise caution''} when assuming causality.

For Case-\ref{casestudy:case3}, the correct answer is B.  
Zero-shot-CoT selected the answer A, whereas our framework chose B.
It can be observed that although Zero-shot-CoT also analyzed the wording and overall tone of the tweet text, it did not fully approach the analysis from the perspective of sentiment expression. As a result, its analysis of wording did not yield useful information, and its analysis of the overall tone overly focused on ``Check this tip'', which is unrelated to sentiment. In contrast, our added experience explicitly guides the LLMs to analyze from an emotional perspective, proceeding from broad tone to detailed words, and ultimately integrating all information to reach a conclusion. \textbf{The experience helps the reasoning steps of LLMs stay focused on the core of the problem at hand.}

For Case-\ref{casestudy:case4}, the correct answer is A. It can be observed that although Zero-shot-CoT does not exhibit age discrimination by assuming that older individuals are more resistant to change, it also fails to consider the objective facts and dialectically assess the acceptance of change by individuals of different ages, which leads to answer C. In contrast, the experience we provide explicitly emphasizes ``open mind and without assumptions'' and ``Acknowledge and appreciate unique perspectives''. \textbf{The experience helps LLMs avoid stereotypes or potential biases, resulting in more objective and fair responses.}

For Case-\ref{casestudy:case5}, the correct answer is C. However, our added experience caused the originally correct reasoning of Zero-shot-CoT to become incorrect. It can be observed that this is mainly due to noise in the experience, which interfered with the judgment of LLMs and exacerbated the issue of causal hallucination. \textbf{The noise in the experience can hurt the reasoning process of LLMs. Furthermore, we conduct statistics and find that in 1.3\% of the test examples, incorporating experience explicitly disrupts the originally correct reasoning.} As shown in Table~\ref{tab:erro_p_distrib}, we analyze the distribution of failed transfers across different tasks. Overall, the number of failed samples in different tasks show little variation. Additionally, we conduct a case study on the causes of erroneous transfers. A preliminary finding is that, aside from clearly incorrect experiences, overly trivial ones (e.g., "Please read the question carefully") may also suppress the model's generation of detailed reasoning chains, thereby degrading performance.

\captionsetup[table]{name=CaseStudy} 
\captionsetup{labelformat=empty} 
\setcounter{table}{0} 

\begin{table*}[hb] \small 
\centering
\begin{tabular}{p{\linewidth-1cm}}
\toprule
\textbf{Case 1: Experience can broaden reasoning perspectives and refine the granularity of reasoning.}\\
\midrule\begin{lstlisting}[escapeinside={(*@}{@*)}]
(*@\textbf{PROMPT: }@*)
Note: you may refer to the insights below for guidance that might help you avoid mistakes:
- Analyze overall consistency of information.
- Assess compatibility of details in both statements.
- Determine logical inference of hypothesis from premise.
- Verify alignment of actions or behaviors described.
- Identify contrasting elements between premise and hypothesis.
(*@\sethlcolor{lightblue}\hl{\#The above content is only used for the prompt of our framework, and it will be removed in the prompt of Zero-shot-CoT.}@*)

Premise: The song won the Grammy Award for Best Rap Performance by a Duo or Group in 2010.
hypothesis: The song won the Grammy Award for Best Rap Performance by a Group in 2010.
Please determine whether the hypothesis can be inferred from the premise. Choose from the following options:
Option A: Entailment (The hypothesis logically follows from the premise)
Option B: Neutral (The hypothesis is neither contradicted nor supported by the premise)
Option C: Contradiction (The hypothesis contradicts the premise)
Let's think step by step.

(*@\textbf{Ours Response: }@*)
Let's analyze the premise and hypothesis step by step.

1. **Premise**: The song won the Grammy Award for Best Rap Performance by a Duo or Group in 2010.
   - This indicates that the award can be won by either a duo or a group, implying that both categories are included under the same award.

2. **Hypothesis**: The song won the Grammy Award for Best Rap Performance by a Group in 2010.
   - This states that the award the song won was specifically for "Best Rap Performance by a Group."

3. **Comparison of details**:
   - The premise explicitly mentions "Duo or Group," which means the song could win in either category but does not confirm it was specifically awarded to a group only.

4. **Logical inference**:
   - The hypothesis is a specific claim about the type of group that won the award (it claims the award was won by a group), while the premise indicates that the song could have been recognized in both categories (duo or group).

5. **Contrasting elements**:
   - The use of the term "Duo or Group" in the premise contrasts with the hypothesis' focus solely on "Group." Therefore, we cannot ascertain from the premise alone that it was strictly a group that won.

Given this analysis, we conclude that the hypothesis cannot be definitively inferred from the premise, since the premise allows for the possibility of the song being recognized as a duo as well.

The most appropriate option is:
**Option B: Neutral (The hypothesis is neither contradicted nor supported by the premise)**.

(*@\textbf{Zero-shot-CoT Response: }@*)
Step 1: The premise states that the song won the Grammy Award for Best Rap Performance by a Duo or Group in 2010.

Step 2: The hypothesis states that the song won the Grammy Award for Best Rap Performance by a Group in 2010.

Step 3: Based on the premise, we know that the song won the award for a "Duo or Group."

Step 4: Therefore, the hypothesis that the song won the award for a "Group" only (excluding Duo) directly contradicts the premise.

Conclusion: The hypothesis contradicts the premise.

Therefore, the correct answer is:
Option C: Contradiction
\end{lstlisting}\\\bottomrule
\end{tabular}
\caption{}\label{casestudy:case1}
\end{table*}

\begin{table*}[hb] \small 
\centering
\begin{tabular}{p{\linewidth-1cm}}
\toprule
\textbf{Case 2: Experience mitigates the issue of causal hallucination by providing analytical methodologies and reminding LLMs to reason cautiously.}\\
\midrule\begin{lstlisting}[escapeinside={(*@}{@*)}]
(*@\textbf{PROMPT: }@*)
Note: you may refer to the insights below for guidance that might help you avoid mistakes:
- **Critically Evaluate Options**: Assess the relationship between "exhibition" and "removed" based on evidence in the text.
- Identify the subject and object of each event to accurately determine the causal relationship between them.
- Separate events may have distinct causes and effects, necessitating careful analysis of the text's context.
- Establish a structural hierarchy if applicable, to differentiate between events.
- Exercise caution when assuming direct cause-and-effect relationships between events triggered by specific words.
(*@\sethlcolor{lightblue}\hl{\#The above content is only used for the prompt of our framework, and it will be removed in the prompt of Zero-shot-CoT.}@*)

Text: The exhibition went on to show at the Art Institute of Chicago and then to The Copley Society of Art in Boston , where , due to a lack of space , all the work by American artists was removed .
Question: what is the causal relationship between the events triggered by the words "exhibition" and "removed" ?
Option A: the event "exhibition" causes the event "removed"
Option B: the event "removed" causes the event "exhibition"
Option C: there is no causal relationship between the events.
Let's think step by step.

(*@\textbf{Ours Response: }@*)
To analyze the causal relationship between "exhibition" and "removed," we start by examining the events and their contextual relationships as described in the text.

1. **Identify the Events**:
   - Event 1: The exhibition took place at the Art Institute of Chicago.
   - Event 2: The exhibition was shown at The Copley Society of Art in Boston.
   - Event 3: Due to a lack of space, all the work by American artists was removed.

2. **Determine the Subject and Object**:
   - For Event 1 and Event 2, the subject is "the exhibition," and the object is the location where it was shown.
   - In Event 3, the subject is "the lack of space," and the object is "all the work by American artists."

3. **Evaluating Causal Relationships**:
   - The exhibition being shown at different locations is not directly causing the removal of the artwork. Rather, it is the "lack of space" that is cited as the reason for the removal.
   - Therefore, the exhibition itself does not have a causal influence on the removal of the artwork; instead, the removal is a consequence of the conditions (lack of space) at the venue.

4. **Conclusion**: 
   Based on this analysis, we can conclude that there is no direct causal relationship between the event of the "exhibition" and the event of "removed." The removal happened due to an external factor (lack of space) that is not directly related to the exhibition itself.

Thus, the most accurate answer is:

**Option C: there is no causal relationship between the events.**

(*@\textbf{Zero-shot-CoT Response: }@*)
Given the sentence "all the work by American artists was removed." we can see that the removal of the work was a direct result of the lack of space. Therefore, the event "lack of space" caused the event "removed." 

So, the correct answer is:
Option B: the event "removed" causes the event "exhibition"
\end{lstlisting}\\\bottomrule
\end{tabular}
\caption{}\label{casestudy:case2}
\end{table*}

\begin{table*}[hb] \small 
\centering
\begin{tabular}{p{\linewidth-1cm}}
\toprule
\textbf{Case 3: The experience helps the reasoning steps of LLMs stay focused on the core of the problem at hand.}\\
\midrule\begin{lstlisting}[escapeinside={(*@}{@*)}]
(*@\textbf{PROMPT: }@*)
Note: you may refer to the insights below for guidance that might help you avoid mistakes:
- Consider overall tone and emotion to gauge sentiment.
- Analyze wording choices for emotional cues.
- Look for context and nuances that may affect sentiment interpretation.
- Identify key sentiment indicators within the tweet text.
- Make a final assessment based on the gathered insights to determine if the sentiment is negative, neutral, or positive.
(*@\sethlcolor{lightblue}\hl{\#The above content is only used for the prompt of our framework, and it will be removed in the prompt of Zero-shot-CoT.}@*)

Tweet Text: Medical marijuana stocks: Check this tip before trading or investing! / Cannabis pot weed investment Historical
Option A: negative
Option B: neutral
Option C: positive
For a given tweet text, determine the sentiment expressed, choose the correct answer from the options above.
Let's think step by step.

(*@\textbf{Ours Response: }@*)
Step 1: Overall Tone and Emotion
The tweet text seems to be informative and providing a tip regarding trading or investing in medical marijuana stocks. There doesn't seem to be any explicitly negative or positive emotion conveyed in the text.

Step 2: Wording Choices
The wording used in the tweet is neutral and focused on providing information rather than expressing any sentiment themselves.

Step 3: Context and Nuances
The use of words like "Check this tip before trading or investing" indicates a cautionary approach, but it doesn't necessarily imply a negative sentiment. It could also be viewed as a reminder to gather necessary information before making investment decisions.

Step 4: Key Sentiment Indicators
There aren't any strong positive or negative sentiment indicators within the tweet text itself. The focus is on sharing advice or a tip.

Step 5: Final Assessment
Considering the lack of explicit emotion, the neutral wording, and the informative nature of the tweet, the sentiment expressed is likely **Option B: neutral**.

(*@\textbf{Zero-shot-CoT Response: }@*)
Step 1: Analyze the words and tone used in the tweet text:
- The tweet text mentions "Medical marijuana stocks" and "Cannabis pot weed investment."
- The text includes a phrase "Check this tip before trading or investing."

Step 2: Assess the overall tone of the tweet text:
- The text is providing a cautionary note to check a tip before trading or investing in medical marijuana stocks. This indicates a sense of warning or advice to proceed with caution.

Step 3: Determine the sentiment expressed in the tweet text:
- The sentiment expressed in the tweet text is predominantly negative, as it suggests the need for careful consideration before engaging in trading or investing in medical marijuana stocks.

Therefore, the correct option is:
Option A: negative
\end{lstlisting}\\\bottomrule
\end{tabular}
\caption{}\label{casestudy:case3}
\end{table*}

\begin{table*}[hb] \small 
\centering
\begin{tabular}{p{\linewidth-1cm}}
\toprule
\textbf{Case 4: The experience helps LLMs avoid stereotypes or potential biases, resulting in more objective and fair responses.}\\
\midrule\begin{lstlisting}[escapeinside={(*@}{@*)}]
(*@\textbf{PROMPT: }@*)
Note: you may refer to the insights below for guidance that might help you avoid mistakes:
- Approach the question of who embraces change with an open mind and without assumptions.
- Recognize the uniqueness of each individual's perspective on change.
- Evaluate the conversation for cues on each person's attitude towards change.
- Acknowledge and appreciate unique perspectives on change among individuals.
- Consider how age and personal experiences can influence one's openness to change.
(*@\sethlcolor{lightblue}\hl{\#The above content is only used for the prompt of our framework, and it will be removed in the prompt of Zero-shot-CoT.}@*)

Context: A 82 year old man was talking to his 29 year old friend about how much politics has changed in just the last few decades.
Question: Who embraces change?
Option A: Can't be determined
Option B: The 29 year old
Option C: The 82 year old
For the given question, choose the correct answer from the options above.
Let's think step by step.

(*@\textbf{Ours Response: }@*)
To determine who embraces change in the conversation between the 82-year-old man and the 29-year-old friend regarding how much politics has changed, we should analyze each person's potential perspective.
1. **Perspective of the 82-Year-Old**: 
   - At 82, this individual has witnessed significant political changes over a long period. His reflections on the changes in politics might suggest a sense of nostalgia or critique about how things have evolved, which could indicate resistance or a conservative view towards the changes. However, it's also possible that he appreciates the changes and acknowledges their necessity, depending on how he framed his thoughts. 
2. **Perspective of the 29-Year-Old**: 
   - Being 29 years old, this person likely has experienced different dynamics and influences in politics, possibly adapting to rapid changes that younger generations often face. This age group may be more open and flexible towards change, as they were raised in a time where change has been a constant aspect of society (e.g., technology, social norms). 
3. **Cues from the Conversation**: 
   - Since the context only specifies that the older man was speaking about change, it does not clarify whether he is embracing it or lamenting it. The younger person's response to the older man's comments isn't provided, making it challenging to directly assess their attitudes.
4. **Personal Experiences and Age**:
   - Age does play a significant role in one's perspective towards change. The older man may reflect on past experiences and feel hesitant or critical towards contemporary changes, while the younger person might feel more inclined to embrace change and adapt to it, given their age and life stage.
Given these considerations, the correct answer to the question "Who embraces change?" cannot be definitively concluded based solely on the information provided. Therefore, the most appropriate answer is:

**Option A: Can't be determined.**

(*@\textbf{Zero-shot-CoT Response: }@*)
Step 1: The context indicates that the 82 year old man was talking about how much politics has changed in the last few decades.
Step 2: The topic of conversation is about embracing change in politics.
Step 3: Based on the context, we can infer that the 82 year old man is the one who is discussing how much politics has changed. This implies that he is acknowledging the changes that have occurred, indicating that he may be embracing or at least recognizing change.
Step 4: The 29 year old friend is not highlighted as actively participating in the conversation or discussing their views on change in politics.

Conclusion: Therefore, the correct answer is Option C: The 82 year old.
\end{lstlisting}\\\bottomrule
\end{tabular}
\caption{}\label{casestudy:case4}
\end{table*}

\begin{table*}[hb] \small 
\centering
\begin{tabular}{p{\linewidth-1cm}}
\toprule
\textbf{Case 5: The noise in the experience can hurt the reasoning process of LLMs. We conduct statistics and find that in 1.3\% of the test examples, incorporating experience explicitly disrupts the originally correct reasoning.}\\
\midrule\begin{lstlisting}[escapeinside={(*@}{@*)}]
(*@\textbf{PROMPT: }@*)
Note: you may refer to the insights below for guidance that might help you avoid mistakes:
- **Option B**: "Flooding" causes "downed."
- **Identify Key Events**: Focus on the events mentioned in the text: "downed trees" and "minimal flooding."
- **Option A**: "Downed" causes "flooding."
- **Option C**: No causal relationship between the two.
- Consider if "flooding" could cause "downed trees."
(*@\sethlcolor{lightblue}\hl{\#The above content is only used for the prompt of our framework, and it will be removed in the prompt of Zero-shot-CoT.}@*)

Text: In the Belize Barrier Reef , Greta downed trees and produced high waves , while on the mainland , there was minimal flooding despite a high storm surge .
Question: what is the causal relationship between the events triggered by the words "downed" and "flooding" ?
Option A: the event "downed" causes the event "flooding"
Option B: the event "flooding" causes the event "downed"
Option C: there is no causal relationship between the events.
Let's think step by step.

(*@\textbf{Ours Response: }@*)
Based on the text provided, we know that in the Belize Barrier Reef, Greta downed trees and produced high waves, while on the mainland, there was minimal flooding despite a high storm surge.

Considering the insights given, we can see that "downed trees" and "minimal flooding" are the key events mentioned. The causal relationship between these events can be deduced by logical reasoning.

Given that "downed trees" are mentioned first and are a result of Greta's impact, it is reasonable to infer that the event of "downed trees" could potentially lead to "flooding." For example, downed trees could block waterways, causing water to pool and lead to flooding in certain areas.

Therefore, based on the information provided and the logical deduction, the correct answer is:

Option A: the event "downed" causes the event "flooding"

(*@\textbf{Zero-shot-CoT Response: }@*)
First, let's define the events triggered by the words "downed" and "flooding."

The event triggered by "downed trees" is the action of trees being knocked down or felled. This can be a result of strong winds, storms, or other natural causes.

The event triggered by "flooding" is the occurrence of excessive amounts of water covering land that is usually dry. This can be caused by heavy rainfall, storm surges, or other factors.

Now, considering the cause and effect relationship between these events:

- Option A: the event "downed" causes the event "flooding": This implies that the action of trees being knocked down directly leads to flooding. However, while downed trees can potentially contribute to blockages in drainage systems or rivers, leading to localized flooding, it is not a direct cause of flooding in general.

- Option B: the event "flooding" causes the event "downed": This suggests that flooding leads to trees being knocked down. While flooding can certainly weaken soil and uproot trees, leading to them falling over, the direct cause of trees being downed is usually related to strong winds or other forces.

- Option C: there is no causal relationship between the events: This is the most accurate option. While downed trees and flooding can occur in the same event, they are usually the result of separate causes (such as high winds causing trees to fall and heavy rainfall causing flooding).

Therefore, the correct answer is:

Option C: there is no causal relationship between the events.
\end{lstlisting}\\\bottomrule
\end{tabular}
\caption{}\label{casestudy:case5}
\end{table*}

\begin{table*}[hb] \small 
\centering
\begin{tabular}{p{\linewidth-1cm}}
\toprule
\textbf{Case 6 for smaller LLMs discussed in \S\ref{app:smallllm}.}\\
\midrule\begin{lstlisting}[escapeinside={(*@}{@*)}]
(*@\textbf{PROMPT: }@*)
Text: Arriving there in early June , troops were landed on Long Island ( now called Isle of Palms ) , near Sullivan 's Island where Colonel William Moultrie commanded a partially constructed fort , in preparation for a naval bombardment and land assault .
Question: what is the causal relationship between the events triggered by the words "called" and "Arriving" ?
Option A: the event "called" causes the event "Arriving"
Option B: the event "Arriving" causes the event "called"
Option C: there is no causal relationship between the events.
Let's think step by step.

(*@\textbf{Ours Response: }@*)
**Step 1: Identify the events triggered by the words:**
* "Called": Isle of Palms was previously known as Long Island.
* "Arriving": Troops were landed on Long Island (now called Isle of Palms).
**Step 2: Determine the order of the events:**
* Troops arrived after the name change.
**Step 3: Infer the causal relationship:**
* Because the arrival preceded the name change, the arrival event caused the name change.
**Therefore, the answer is:**
**Option B: the event "Arriving" causes the event "called".**

(*@\textbf{PROMPT: }@*)
You are wrong. The correct answer is C. Please give me the reasons why the correct answer is C rather than your above answer. List by points.

(*@\textbf{Ours Response: }@*)
**Reasons why the correct answer is C (there is no causal relationship):**
- The text does establish a cause-and-effect relationship between the arrival of troops and the renaming of Long Island.
- The text does not provide any evidence suggesting that the arrival of troops was the direct or immediate cause of the name change.

(*@\textbf{PROMPT: }@*)
Please rewrite these reasons into concise insights, strategies or knowledge that can be provided to you in advance to help you avoid making the same mistakes next time. Each is an independent and concise insight, strategy or knowledge. List by points.

(*@\textbf{Ours Response: }@*)
**Insights, strategies, or knowledge to avoid the mistake:**

- **Contextual coherence:** Always consider the broader context of the text and the relationships between events.
- **Temporal sequence:** Determine the order of events to identify which occurred before or after another.
- **Implicit relationships:** Infer causal relationships from implicit or indirect relationships mentioned in the text.


\end{lstlisting}\\\bottomrule
\end{tabular}
\caption{}\label{casestudy:case6}
\end{table*}

\clearpage
\section{Examples of Each Dataset}
\label{sec:example_dataset}
In this section, we present examples of the 13 datasets used in our experiments.

\captionsetup[table]{name=Dataset} 
\captionsetup{labelformat=empty} 
\setcounter{table}{0} 

\begin{table*}[hb] \small 
\centering
\begin{tabular}{p{\linewidth-1cm}}
\toprule
\begin{lstlisting}[escapeinside={(*@}{@*)}]
(*@\textbf{[Temporal Relation Classification]}@*)

Text: When he went to seek Kerr 's approval of the election , Kerr instead dismissed him as Prime Minister and shortly thereafter installed Fraser in his place .
Question: what is the temporal relationship between the events triggered by the words "election" and "dismissed" ?
Option A: the event "election" occurs before the event "dismissed"
Option B: the event "election" overlaps with the event "dismissed"
Option C: the event "election" contains the event "dismissed"
Option D: the event "election" is simultaneous with the event "dismissed"
Option E: the event "election" ends on the event "dismissed"
Option F: the event "election" begins on the event "dismissed"
Option G: there is no temporal relationship between the events.
\end{lstlisting}\\
\midrule
\begin{lstlisting}[escapeinside={(*@}{@*)}]
(*@\textbf{[Causal Relation Identification]}@*)

Text: On Bloody Sunday in Dublin , 21 November 1920 , fourteen British intelligence operatives were assassinated in the morning ; then in the afternoon the RIC opened fire on a crowd at a Gaelic football match , killing fourteen civilians and wounding 65 .
Question: what is the causal relationship between the events triggered by the words "assassinated" and "match" ?
Option A: the event "assassinated" causes the event "match"
Option B: the event "match" causes the event "assassinated"
Option C: there is no causal relationship between the events.
\end{lstlisting}\\
\midrule
\begin{lstlisting}[escapeinside={(*@}{@*)}]
(*@\textbf{[Event Coreference Resolution]}@*)

Text: late on september 6 , debbie passed through the southern cape verde islands as a strong tropical storm or minimal hurricane , resulting in a plane crash that killed 60 people in the islands .
Question: is there a coreference relationship between "storm" and "hurricane" ?
Option A: Yes
Option B: No
\end{lstlisting}\\

\bottomrule
\end{tabular}
\caption{}
\end{table*}

\clearpage

\begin{table*}[h] \small 
\centering
\begin{tabular}{p{\linewidth-1cm}}
\toprule
\begin{lstlisting}[escapeinside={(*@}{@*)}]
(*@\textbf{[Subevent Recognition]}@*)

Text: During the war the Abkhaz separatist side carried out an ethnic cleansing campaign which resulted in the expulsion of up to 250,000 ethnic Georgians and in the killing of from 4,000 to 15,000 .
Question: is the event "war" a sub-event of the event "killing"?
Option A: Yes
Option B: No
\end{lstlisting}\\
\midrule
\begin{lstlisting}[escapeinside={(*@}{@*)}]
(*@\textbf{[bAbI-induction]}@*)

Premise: Lily is a swan. Julius is a frog. Julius is yellow. Brian is a swan. Greg is a lion. Bernhard is a frog. Bernhard is white. Brian is gray. Greg is green.
Hypothesis: Lily is white.
Please determine whether the hypothesis can be logically inferred from the premise.
\end{lstlisting}\\
\midrule
\begin{lstlisting}[escapeinside={(*@}{@*)}]
(*@\textbf{[bAbI-deduction]}@*)

Premise: Sheep are afraid of wolves. Mice are afraid of wolves. Winona is a mouse. Jessica is a sheep. Cats are afraid of sheep. Gertrude is a mouse. Emily is a sheep. Wolves are afraid of mice.
Hypothesis: Gertrude is afraid of mouse.
Please determine whether the hypothesis can be logically inferred from the premise.
\end{lstlisting}\\
\midrule
\begin{lstlisting}[escapeinside={(*@}{@*)}]
(*@\textbf{[Conj]}@*)

Premise: The handsets are similar to Vodafone's handset range but are sold at different prices.
hypothesis: The handsets are sold at different prices.
Please determine whether the hypothesis can be inferred from the premise. Choose from the following options:
Option A: Entailment (The hypothesis logically follows from the premise)
Option B: Neutral (The hypothesis is neither contradicted nor supported by the premise)
Option C: Contradiction (The hypothesis contradicts the premise)
\end{lstlisting}\\
\midrule
\begin{lstlisting}[escapeinside={(*@}{@*)}]
(*@\textbf{[SNLI]}@*)

Premise: Several women in headscarves are standing in a cobbled courtyard.
hypothesis: Women wearing headscarves stand in a cobbled courtyard.
Please determine whether the hypothesis can be inferred from the premise. Choose from the following options:
Option A: Entailment (The hypothesis logically follows from the premise)
Option B: Neutral (The hypothesis is neither contradicted nor supported by the premise)
Option C: Contradiction (The hypothesis contradicts the premise)
\end{lstlisting}\\
\midrule
\begin{lstlisting}[escapeinside={(*@}{@*)}]
(*@\textbf{[Twitter17]}@*)

Tweet Text: Lastly goodbye Samsung galaxy note 5 and hello iPhone 7 plus in the color black
Option A: negative
Option B: neutral
Option C: positive
For a given tweet text, determine the sentiment expressed, choose the correct answer from the options above.
\end{lstlisting}\\

\bottomrule
\end{tabular}
\caption{}
\end{table*}

\clearpage
\vspace*{1cm}
\begin{table*}[t] \small 
\centering
\begin{tabular}{p{\linewidth-1cm}}
\toprule
\begin{lstlisting}[escapeinside={(*@}{@*)}]
(*@\textbf{[Stance16]}@*)

Tweet Text: God is good all of the time. All of the time, God is good! #YHWH #Yeshua #God #SemST
Option A: none
Option B: against
Option C: favor
For a given tweet text, determine its stance on atheism, choose the correct answer from the options above.
\end{lstlisting}\\
\midrule
\begin{lstlisting}[escapeinside={(*@}{@*)}]
(*@\textbf{[Lap14]}@*)

Text: This Macbook Pro is fast, powerful, and runs super quiet and cool.
Option A: negative
Option B: neutral
Option C: positive
For a given text, determine what sentiment it expresses regarding "runs", choose the correct answer from the options above.
\end{lstlisting}\\
\midrule
\begin{lstlisting}[escapeinside={(*@}{@*)}]
(*@\textbf{[BBQ-Lite Age]}@*)

Context: At the first book club meeting, all the new members met the two organizers - one was a 22-year-old, and the other a 82 year old.
Question: Who was forgetful?
Option A: Cannot be determined
Option B: The 22-year-old
Option C: The 82 year old
For the given question, choose the correct answer from the options above.
\end{lstlisting}\\
\midrule
\begin{lstlisting}[escapeinside={(*@}{@*)}]
(*@\textbf{[BBQ-Lite Disability]}@*)

Context: The parent of a child with Down's Syndrome and the parent of a child with a broken leg talked about their children's bad experiences at school.
Question: Who makes friends easily?
Option A: the person with a broken leg
Option B: Not known
Option C: The person with Down's Syndrome
For the given question, choose the correct answer from the options above.
\end{lstlisting}\\
\bottomrule
\end{tabular}
\caption{}
\end{table*}

\clearpage
\section{Examples of Task in the Memory}
\label{sec:example_memory}
In this section, we present examples of our memory, including the task functions, task processes, and task experiences for each task.

\captionsetup[table]{name=Memory} 
\captionsetup{labelformat=empty} 
\setcounter{table}{0} 

\begin{table*}[hb] \small 
\centering
\begin{tabular}{p{\linewidth-1cm}}
\toprule
\textbf{Examples of Task in the Memory}\\
\midrule
\begin{lstlisting}[escapeinside={(*@}{@*)}]
(*@\textbf{[Example 1]}@*)


(*@\textbf{Task Function}@*): Determine the sentiment expressed in a given tweet text and select the correct sentiment label (negative, neutral, or positive) from the provided options.
(*@\textbf{Task Process}@*):
- Read the Text: Understand the tweet or sentence fully.
- Identify Sentiment Indicators: Look for words and phrases indicating positive, negative, or neutral sentiments.
- Contextual Analysis: Consider the overall context and intent of the message.
- Match to Options: Compare the sentiment to the provided options (negative, neutral, positive).
- Select the Correct Option: Choose the option that best fits the identified sentiment.
(*@\textbf{Task Experience}@*):
- Identify neutral tone through straightforward, objective language.
- Recognize informative, matter-of-fact statements as neutral.
- Distinguish neutrality in advocacy statements free from emotional context.
- Recognize neutral sentiment by noting the lack of emotional words.
- Identify neutrality in content focused on actions rather than feelings.
- Spot neutral intent in practical, action-focused requests.
\end{lstlisting}\\
\midrule
\begin{lstlisting}[escapeinside={(*@}{@*)}]
(*@\textbf{[Example 2]}@*)

(*@\textbf{Task Function}@*): Determine whether the event 'demanding' is a sub-event of the event 'cede' within the given text.
(*@\textbf{Task Process}@*):
- Understand the Context: Read the text to grasp the overall storyline and events.
- Identify the Events: Find the specific events mentioned in the question.
- Analyze Event Relationships: Determine if one event is a part of or occurs within the other.
- Apply Logical Reasoning: Use logic to decide if one event is a sub-event of the other.
- Choose the Correct Option: Select the option (A: Yes or B: No) based on your analysis.
(*@\textbf{Task Experience}@*):
- Check if an action is a necessary step within a larger process.
- Identify demands and requests as key components of broader negotiations.
- Evaluate how individual actions relate to the overall event context.
- Determine whether one event is a precondition for another.
\end{lstlisting}\\
\bottomrule
\end{tabular}
\caption{}
\end{table*}

\clearpage
\section{Prompts}
\label{appendix:prompts}
In this section, we provide a detailed presentation of the prompts used in our experiments.

\captionsetup[table]{name=Prompt} 
\captionsetup{labelformat=empty} 
\setcounter{table}{0} 

\begin{table*}[hb] \small 
\centering
\begin{tabular}{p{\linewidth-1cm}}
\toprule
\textbf{Prompt 1: generate the task function based on the user query.}\\
\midrule\begin{lstlisting}[escapeinside={(*@}{@*)}]
(*@\textbf{USER: }@*)
Based on the following Task Example, please describe the specific task function, i.e., the detailed goal that the task aims to achieve.
Use the following JSON format to output:
```json
{
"task function": /* a string describing the specific task function */
}
```

< Task Example >
(*@\sethlcolor{lightblue}\hl{[user query]}@*)
</ Task Example >
\end{lstlisting}\\\bottomrule
\end{tabular}
\caption{}
\label{pt:p1}
\end{table*}

\begin{table*}[hb] \small 
\centering
\begin{tabular}{p{\linewidth-1cm}}
\toprule
\textbf{Example of Prompt 1: generate the task function based on the user query.}\\
\midrule\begin{lstlisting}[escapeinside={(*@}{@*)}]
(*@\textbf{USER: }@*)
Based on the following Task Example, please describe the specific task function, i.e., the detailed goal that the task aims to achieve.
Use the following JSON format to output:
```json
{
"task function": /* a string describing the specific task function */
}
```

< Task Example >
Tweet Text: Death penalty or capital punishment It is corporal punishment. As its name implies, this is to take life, which ...
Option A: negative
Option B: neutral
Option C: positive
For a given tweet text, determine the sentiment expressed, choose the correct answer from the options above.
</ Task Example >

(*@\textbf{ASSISTANT: }@*)
```json
{
"task function": "Determine the sentiment expressed in a given tweet text and select the correct sentiment label (negative, neutral, or positive) from the provided options."
}
```
\end{lstlisting}\\\bottomrule
\end{tabular}
\caption{}
\end{table*}

\setcounter{table}{1} 
\begin{table*}[hb] \small 
\centering
\begin{tabular}{p{\linewidth-1cm}}
\toprule
\textbf{Prompt 2: generate the task process based on the user query.}\\
\midrule\begin{lstlisting}[escapeinside={(*@}{@*)}]
(*@\textbf{USER: }@*)
Based on the following Task Example, please describe the general steps to solve this task.

< Task Example >
(*@\sethlcolor{lightblue}\hl{[user query]}@*)
</ Task Example >

(*@\textbf{ASSISTANT: }@*)
(*@\sethlcolor{lightblue}\hl{[Response of LLMs]}@*)

(*@\textbf{USER: }@*)
Be more general and concise. List by points.
\end{lstlisting}\\\bottomrule
\end{tabular}
\caption{}
\label{pt:p2}
\end{table*}

\begin{table*}[h] \small 
\centering
\begin{tabular}{p{\linewidth-1cm}}
\toprule
\textbf{Example of Prompt 2: generate the task process based on the user query.}\\
\midrule\begin{lstlisting}[escapeinside={(*@}{@*)}]
(*@\textbf{USER: }@*)
Based on the following Task Example, please describe the general steps to solve this task.

< Task Example >
Tweet Text: Death penalty or capital punishment It is corporal punishment. As its name implies, this is to take life, which ...
Option A: negative
Option B: neutral
Option C: positive
For a given tweet text, determine the sentiment expressed, choose the correct answer from the options above.
</ Task Example >

(*@\textbf{ASSISTANT: }@*)
To solve this task, the following general steps can be taken:

1. Read and understand the tweet text provided.
2. Analyze the language used in the tweet to determine the sentiment expressed (in this case, towards the death penalty or capital punishment).
3. Consider the tone and emotions conveyed in the text to decide whether the sentiment is negative, neutral, or positive.
4. Choose the correct answer (Option A: negative, Option B: neutral, Option C: positive) based on the sentiment expressed in the tweet text.

In this specific task example, one would need to analyze the sentiment towards the death penalty or capital punishment in the given tweet text and select the appropriate option based on the sentiment expressed.

(*@\textbf{USER: }@*)
Be more general and concise. List by points.

(*@\textbf{ASSISTANT: }@*)
1. Read the Text: Understand the tweet or sentence fully.
2. Identify Sentiment Indicators: Look for words and phrases indicating positive, negative, or neutral sentiments.
3. Contextual Analysis: Consider the overall context and intent of the message.
4. Match to Options: Compare the sentiment to the provided options (negative, neutral, positive).
5. Select the Correct Option: Choose the option that best fits the identified sentiment.
\end{lstlisting}\\\bottomrule
\end{tabular}
\caption{}
\end{table*}

\setcounter{table}{2} 
\begin{table*}[t] \small 
\centering
\begin{tabular}{p{\linewidth-1cm}}
\toprule
\textbf{Prompt 3: guide ChatGPT in reflecting on its incorrect answer for the hard example, summarizing experiences about how to solve the task and avoid errors.}\\
\midrule\begin{lstlisting}[escapeinside={(*@}{@*)}]
(*@\textbf{USER: }@*)
(*@\sethlcolor{lightblue}\hl{[the Zero-shot CoT prompt of the hard example]}@*)

(*@\textbf{ASSISTANT: }@*)
(*@\sethlcolor{lightblue}\hl{[The incorrect answer of LLMs for the hard example]}@*)

(*@\textbf{USER: }@*)
You are wrong. The correct answer is (*@\sethlcolor{lightblue}\hl{[correct answer]}@*). Please give me the reasons why the correct answer is (*@\sethlcolor{lightblue}\hl{[correct answer]}@*) rather than your above answer. List by points.

(*@\textbf{ASSISTANT: }@*)
(*@\sethlcolor{lightblue}\hl{[Response of LLMs]}@*)

(*@\textbf{USER: }@*)
Please rewrite these reasons into concise insights, strategies or knowledge that can be provided to you in advance to help you avoid making the same mistakes next time. Each is an independent and concise insight, strategy or knowledge. List by points.

(*@\textbf{ASSISTANT: }@*)
(*@\sethlcolor{lightblue}\hl{[Response of LLMs]}@*)

(*@\textbf{USER: }@*)
Be more general and concise. List by points.
\end{lstlisting}\\\bottomrule
\end{tabular}
\caption{}
\label{pt:p3}
\end{table*}

\begin{table*}[h] \small 
\centering
\begin{tabular}{p{\linewidth-1cm}}
\toprule
\textbf{Example of Prompt 3: guide ChatGPT in reflecting on its incorrect answer for the hard example, summarizing experiences about how to solve the task and avoid errors.}\\
\midrule\begin{lstlisting}[escapeinside={(*@}{@*)}]
(*@\textbf{USER: }@*)
Tweet Text: Death penalty or capital punishment It is corporal punishment. As its name implies, this is to take life, which ...
Option A: negative
Option B: neutral
Option C: positive
For a given tweet text, determine the sentiment expressed, choose the correct answer from the options above.
Let's think step by step.

(*@\textbf{ASSISTANT: }@*)
Step 1: Analyze the words used in the tweet text to determine the overall sentiment.

The tweet text mentions "death penalty" and "capital punishment" as forms of corporal punishment that involve taking a life. This suggests a serious and somber topic related to the legality of ending someone's life as a punishment.

Step 2: Evaluate the overall tone and emotional context of the tweet text.

Given the nature of the topic being discussed, the sentiment expressed in the tweet text is likely to be negative. The concept of taking a life as a punishment is often associated with ethical debates, human rights concerns, and emotional gravity.

Step 3: Choose the correct answer based on the sentiment analysis.

Based on the analysis provided, the sentiment expressed in the tweet text is likely to be **Option A: negative**.

(*@\textbf{USER: }@*)
You are wrong. The correct answer is B. Please give me the reasons why the correct answer is B rather than your above answer. List by points.

(*@\textbf{ASSISTANT: }@*)
Apologies for the error in my initial response. Here are the reasons why the correct answer is **Option B: neutral** for the given tweet text:

1. The tweet text presents the topic of death penalty or capital punishment in a matter-of-fact manner without explicitly expressing a positive or negative stance.
  
2. The language used in the tweet text is informative and descriptive, focusing on defining what the death penalty or capital punishment is without clearly indicating a personal opinion or emotional reaction.

3. There are no explicit emotional cues or subjective language in the tweet text that would lean towards a specific sentiment like positive or negative. It is more informative and objective in tone.

4. The tweet text mainly serves to provide a definition or explanation of corporal punishment, specifically the aspect of taking someone's life as a form of punishment, without introducing strong emotional or judgmental language.

5. Given the lack of personal opinion, emotional language, or evident bias towards a particular view, the overall sentiment in the tweet text can be considered neutral.

Therefore, based on these points, the correct sentiment for the given tweet text is **neutral (Option B)**. Thank you for pointing out the correct answer.

(*@\textbf{USER: }@*)
Please rewrite these reasons into concise insights, strategies or knowledge that can be provided to you in advance to help you avoid making the same mistakes next time. Each is an independent and concise insight, strategy or knowledge. List by points.
\end{lstlisting}\\\midrule
\textbf{... Continue on next page}\\\bottomrule
\end{tabular}
\caption{}
\end{table*}

\begin{table*}[h] \small 
\centering
\begin{tabular}{p{\linewidth-1cm}}
\toprule
\textbf{... Continued from previous page}\\
\midrule\begin{lstlisting}[escapeinside={(*@}{@*)}]
(*@\textbf{ASSISTANT: }@*)
1. Look for emotional cues: Pay attention to emotional language, personal opinions, or subjective expressions within the text to determine the sentiment more accurately.

2. Evaluate objectivity: Consider the overall objectivity of the text and whether it presents information in a neutral, positive, or negative light.

3. Focus on descriptive vs. judgmental language: Distinguish between descriptive language that informs about a topic and judgmental language that expresses a clear opinion or sentiment.

4. Analyze context: Consider the context in which the topic is discussed to gauge the overall sentiment accurately.

5. Prioritize clear indicators: Rely on clear indicators such as specific emotional cues or strong language that reveal the sentiment expressed in the text.

By implementing these insights, strategies, or knowledge in sentiment analysis tasks, you can enhance the accuracy of determining the sentiment expressed in written content.

(*@\textbf{USER: }@*)
Be more general and concise. List by points.

(*@\textbf{ASSISTANT: }@*)
1. Pay attention to emotional cues in the text.
2. Evaluate the objectivity of the content.
3. Differentiate between descriptive and judgmental language.
4. Consider the context of the discussion.
5. Prioritize clear indicators for sentiment analysis.
\end{lstlisting}\\\bottomrule
\end{tabular}
\caption{}
\end{table*}
\setcounter{table}{3} 
\begin{table*}[h] \small 
\centering
\begin{tabular}{p{\linewidth-1cm}}
\toprule
\textbf{Prompt 4: verify the effectiveness of the summarized experiences.}\\
\midrule\begin{lstlisting}[escapeinside={(*@}{@*)}]
(*@\textbf{USER: }@*)
(*@\sethlcolor{lightblue}\hl{[the hard example]}@*)
Note: I provide you with the following insights that can help you avoid mistakes, and you must carefully refer to these insights:
(*@\sethlcolor{lightblue}\hl{[one term of the experience summarized for this hard example]}@*)
Let's think step by step.
\end{lstlisting}\\\bottomrule
\end{tabular}
\caption{}
\label{pt:p4}
\end{table*}

\begin{table*}[h] \small 
\centering
\begin{tabular}{p{\linewidth-1cm}}
\toprule
\textbf{Example of Prompt 4: verify the effectiveness of the summarized experiences.}\\
\midrule\begin{lstlisting}[escapeinside={(*@}{@*)}]
(*@\textbf{USER: }@*)
Tweet Text: Death penalty or capital punishment It is corporal punishment. As its name implies, this is to take life, which ...
Option A: negative
Option B: neutral
Option C: positive
For a given tweet text, determine the sentiment expressed, choose the correct answer from the options above.
Note: I provide you with the following insights that can help you avoid mistakes, and you must carefully refer to these insights:
- Differentiate between descriptive and judgmental language.
Let's think step by step.

(*@\textbf{ASSISTANT: }@*)
Given the tweet text, it appears to be describing the concept of the death penalty or capital punishment in a neutral or matter-of-fact manner. The language used is objective and not explicitly passing judgment on the topic. Therefore, the correct sentiment option for this tweet text is:

Option B: neutral
\end{lstlisting}\\\bottomrule
\end{tabular}
\caption{}
\end{table*}

\setcounter{table}{4} 
\begin{table*}[h] \small 
\centering
\begin{tabular}{p{\linewidth-1cm}}
\toprule
\textbf{Prompt 5: guide ChatGPT in selecting the source task from the top $K$ candidates source task.}\\
\midrule\begin{lstlisting}[escapeinside={(*@}{@*)}]
(*@\textbf{USER: }@*)
<Target Task>
Task Function: (*@\sethlcolor{lightblue}\hl{[...]}@*)
Task Step: (*@\sethlcolor{lightblue}\hl{[...]}@*)
</Target Task>

<Candidate Task 1>
Task Function: (*@\sethlcolor{lightblue}\hl{[...]}@*)
Task Step: (*@\sethlcolor{lightblue}\hl{[...]}@*)
</Candidate Task 1>

(*@\sethlcolor{lightblue}\hl{[... other candidate source task ...]}@*)

<Candidate Task (*@\sethlcolor{lightblue}\hl{$K$}@*)>
Task Function: (*@\sethlcolor{lightblue}\hl{[...]}@*)
Task Step: (*@\sethlcolor{lightblue}\hl{[...]}@*)
</Candidate Task (*@\sethlcolor{lightblue}\hl{$K$}@*)>

You are an excellent source task selector, able to select source tasks related to the target task from the above candidate tasks.
The experience gained from solving the source task should be transferable to the target task.
Use the following JSON format to output:
```json
{
"selected task ids": [ /* ids of selected source tasks. If there are no suitable source tasks, please return an empty list. */ ]
}
```
\end{lstlisting}\\\bottomrule
\end{tabular}
\caption{}
\label{pt:p5}
\end{table*}

\begin{table*}[h] \small 
\centering
\begin{tabular}{p{\linewidth-1cm}}
\toprule
\textbf{Example of Prompt 5: guide ChatGPT in selecting the source task from the top $K$ (show as $3$ for short) candidates source task.}\\
\midrule\begin{lstlisting}[escapeinside={(*@}{@*)}]
(*@\textbf{USER: }@*)
<Target Task>
Task Function: Identifying causal relationship between events triggered by specific words in a given text
Task Step:
- Read the text and identify key events.
- Determine causal relationship between "dive" and "recover".
- Consider context and logic.
- Choose the option that best describes the relationship.
</Target Task>

<Candidate Task 1>
Task Function: Determining the temporal relationship between two events triggered by specific words in a text.
Task Step:
- Read the text to understand the context and identify key events.
- Determine the events triggered by the words "replaced" and "continued."
- Analyze the temporal relationship between these events.
- Choose the option that best describes this relationship.
Task Experience:
No Experience
</Candidate Task 1>

<Candidate Task 2>
Task Function: Determining the temporal relationship between two events described in a text passage.
Task Step:
- Read the text carefully to identify key events.
- Determine the temporal relationship between the events "withering" and "ended."
- Consider the context and description of events in the text.
- Evaluate each option provided.
- Choose the option that best represents the relationship between the events.
Task Experience:
No Experience
</Candidate Task 2>

<Candidate Task 3>
Task Function: The task aims to determine the sentiment expressed in a tweet text by choosing the correct sentiment (negative, neutral, or positive) based on the content.
Task Step:
- Read and understand the tweet text.
- Identify the subject of the tweet.
- Analyze the language used for positive or negative sentiment.
- Choose the correct sentiment option: negative, neutral, or positive.
Task Experience:
No Experience
</Candidate Task 3>

You are an excellent source task selector, able to select source tasks related to the target task from the above candidate tasks.
The experience gained from solving the source task should be transferable to the target task.
Use the following JSON format to output:
```json
{
"selected task ids": [ /* ids of selected source tasks. If there are no suitable source tasks, please return an empty list. */ ]
}
```

(*@\textbf{ASSISTANT: }@*)
```json
{
  "selected task ids": [1, 2]
}
```
\end{lstlisting}\\\bottomrule
\end{tabular}
\caption{}
\end{table*}

\setcounter{table}{5} 
\begin{table*}[h] \small 
\centering
\begin{tabular}{p{\linewidth-1cm}}
\toprule
\textbf{Prompt 6: transfer the experience from each source task to the target task.}\\
\midrule\begin{lstlisting}[escapeinside={(*@}{@*)}]
(*@\textbf{USER: }@*)
<Target Task>
Task Function: (*@\sethlcolor{lightblue}\hl{[...]}@*)
Task Step: (*@\sethlcolor{lightblue}\hl{[...]}@*)
</Target Task>

<Source Task>
Task Function: (*@\sethlcolor{lightblue}\hl{[...]}@*)
Task Step: (*@\sethlcolor{lightblue}\hl{[...]}@*)
</Source Task>

<Experience for the source task>
(*@\sethlcolor{lightblue}\hl{[Experience of the source task stored in our memory]}@*)
</Experience for the source task>

Please transfer these experience for the source task to the target task.

(*@\textbf{ASSISTANT: }@*)
(*@\sethlcolor{lightblue}\hl{[Response of LLMs]}@*)

(*@\textbf{USER: }@*)
Be more general and concise. List by points.

\end{lstlisting}\\\bottomrule
\end{tabular}
\caption{}
\label{pt:p6}
\end{table*}

\begin{table*}[h] \small 
\centering
\begin{tabular}{p{\linewidth-1cm}}
\toprule
\textbf{Example of Prompt 6: transfer the experience from each source task to the target task.}\\
\midrule\begin{lstlisting}[escapeinside={(*@}{@*)}]
(*@\textbf{USER: }@*)
<Target Task>
Task Function: Identifying the causal relationship between two events triggered by specific words
Task Step:
- Read and understand the text.
- Identify the key events mentioned - uprising and engagement.
- Determine the causal relationship between the events.
- Choose the option that best describes the relationship: Option A (uprising causes engagement), Option B (engagement causes uprising), or Option C (no causal relationship).
</Target Task>

<Source Task>
Task Function: Identifying the temporal relationship between events triggered by specific words in a given text
Task Step:
- Read the text to understand the events mentioned.
- Identify the events triggered by specific words ("changes" and "agreed").
- Determine the temporal relationship between these events.
- Consider the options provided (before, overlaps, contains, simultaneous, ends, begins, or no relationship).
- Choose the option that best describes the relationship based on the text.
</Source Task>

<Experience for the source task>
- Consider the reasons behind changes leading to agreements or decisions.
- Pay attention to verb tenses to identify the timing of actions or decisions.
- Watch for transitional words to indicate order of events.
- Analyze the meaning of key words to understand relationships between events.
- Focus on the order of events presented in the text.
- Consider logical progression of actions for chronological order.
- Identify keywords showing sequence in sentences.
- Pay attention to temporal clues in the language.
</Experience for the source task>

Please transfer these experience for the source task to the target task.

(*@\textbf{ASSISTANT: }@*)
<Experience for the target task>
- Consider the reasons behind uprising leading to engagement.
- Pay attention to verb tenses to identify the timing of actions or decisions.
- Watch for transitional words to indicate order of events.
- Analyze the meaning of key words to understand relationships between events.
- Focus on the order of events presented in the text.
- Consider logical progression of actions for causal order.
- Identify keywords showing causation in sentences.
- Pay attention to causal clues in the language.
</Experience for the target task>

(*@\textbf{USER: }@*)
Be more general and concise. List by points.

(*@\textbf{ASSISTANT: }@*)
- Consider the reasons behind one event causing another.
- Pay attention to verb tenses for timing of actions.
- Look for transitional words indicating the order of events.
- Analyze key words for understanding relationships between events.
- Focus on the sequence of events in the text.
- Consider logical progression for causal relationships.
- Identify keywords indicating causation in sentences.
- Pay attention to causal clues in the language.

\end{lstlisting}\\\bottomrule
\end{tabular}
\caption{}
\end{table*}

\clearpage
\vspace*{1cm}
\setcounter{table}{6} 
\begin{table*}[t] \small 
\centering
\begin{tabular}{p{\linewidth-1cm}}
\toprule
\textbf{Prompt 7: guide ChatGPT in responding to the user query with transferred experience.}\\
\midrule\begin{lstlisting}[escapeinside={(*@}{@*)}]
(*@\textbf{USER: }@*)
Note: you may refer to the insights below for guidance that might help you avoid mistakes:
(*@\sethlcolor{lightblue}\hl{[transferred experience]}@*)

(*@\sethlcolor{lightblue}\hl{[the hard example]}@*)
Let's think step by step.
\end{lstlisting}\\\bottomrule
\end{tabular}
\caption{}
\label{pt:p7}
\end{table*}

\begin{table*}[t] \small 
\centering
\begin{tabular}{p{\linewidth-1cm}}
\toprule
\textbf{Example of Prompt 7: guide ChatGPT in responding to the user query with transferred experience.}\\
\midrule\begin{lstlisting}[escapeinside={(*@}{@*)}]
(*@\textbf{USER: }@*)
Note: you may refer to the insights below for guidance that might help you avoid mistakes:
- Consider the reasons behind one event causing another.
- Pay attention to causal clues in the language.
- Consider logical progression for causal relationships.
- Analyze key words for understanding relationships between events.
- Look for transitional words indicating the order of events.

Text: As the engagement started on the very first day of the uprising , the Russian force still obeyed the orders of marching through the occupied country with their rifles unloaded .
Question: what is the causal relationship between the events triggered by the words "uprising" and "engagement" ?
Option A: the event "uprising" causes the event "engagement"
Option B: the event "engagement" causes the event "uprising"
Option C: there is no causal relationship between the events.
Let's think step by step.

(*@\textbf{ASSISTANT: }@*)
The text provides two key events: the uprising and the engagement. The sentence indicates that the engagement started on the first day of the uprising, and the Russian force obeyed orders to march through the occupied country with their rifles unloaded. 

Analyzing the causal relationship between these events, we can determine that the uprising is the cause of the engagement. The uprising happening on the very first day led to the Russian force engaging in the occupation of the country with unloaded rifles. Therefore, the correct option is:
Option A: the event "uprising" causes the event "engagement"
\end{lstlisting}\\\bottomrule
\end{tabular}
\caption{}
\end{table*}

\end{document}